\documentclass[journal]{IEEEtran}

\usepackage{xcolor,soul,framed} %,caption
\usepackage{makecell}
\colorlet{shadecolor}{yellow}
\usepackage[pdftex]{graphicx}
\graphicspath{{../pdf/}{../jpeg/}}
\DeclareGraphicsExtensions{.pdf,.jpeg,.png}
\usepackage{booktabs}

\usepackage{tikz}
\usetikzlibrary{arrows.meta, shapes.geometric, shapes.misc, positioning, fit,calc}

\usepackage{graphicx}
\usepackage{threeparttable}

\usepackage{algorithm}
\usepackage{algpseudocode}
\usepackage{tabularx}

\usepackage{comment}
\usepackage{url}
\usepackage[cmex10]{amsmath}     
\let\labelindent\relax
\usepackage{enumitem}
\usepackage{amssymb}             % \mathbb, 
\usepackage{array}  
\usepackage{subcaption}

\usepackage{amsthm}

\usepackage{float}

\newtheoremstyle{pkunoindent}% <name>
  {6pt}%   Space above
  {6pt}%   Space below
  {\itshape}% Body font
  {0pt}%   Indent amount  
  {\bfseries}% Theorem head font
  {.}%     Punctuation after theorem head
  {0.5em}% Space after theorem head
  {}%      Theorem head spec

\theoremstyle{pkunoindent}

\newtheorem{proposition}{Proposition}

\usepackage[nocompress]{cite}

\begin{document}
\bstctlcite{IEEEexample:BSTcontrol}
    \title{R\textsuperscript{2}-HGP: A Double-Regularized Gaussian Process for Heterogeneous Transfer Learning}
  \author{Duo Wang,
      Xinming Wang,
      Chao Wang,
      Xiaowei Yue, \IEEEmembership{Senior Member, IEEE},\\
      and Jianguo Wu, \IEEEmembership{Senior Member, IEEE}
      
  \thanks{Duo Wang, and Jianguo Wu are with the Department of Control Science and Systems Engineering, Peking University, Beijing 100871, China (e-mail: j.wu@pku.edu.cn).  (Corresponding author: Jianguo Wu.)
  }
  \thanks{Xinming Wang is with China Mobile Information Technology Co., Ltd., Beijing 100032, China.
  }
   
  \thanks{Chao Wang is with the Department of Industrial and Systems Engineering,
 University of Iowa, Iowa City, IA 52242 USA.}
 
  \thanks{Xiaowei Yue is with the Department of Industrial Engineering, Institute for
Quality and Reliability, Tsinghua University, Beijing 100190, China.}

  }

% \markboth{IEEE TRANSACTIONS ON PATTERN ANALYSIS AND MACHINE INTELLIGENCE
% }{Wang \MakeLowercase{\textit{et al.}}: R\textsuperscript{2}-HGP: A Double-Regularized Heterogeneous Gaussian Process for Transfer Learning}

% ====================================================================
\maketitle

% === ABSTRACT ====================================================================
% =================================================================================
\begin{abstract}
%\boldmath
Multi-output Gaussian process (MGP) models have attracted significant attention for their flexibility and uncertainty-quantification capabilities, and have been widely adopted in multi-source transfer learning scenarios due to their ability to capture inter-task correlations. However, they still face several challenges in transfer learning. First, the input spaces of the source and target domains are often heterogeneous, which makes direct knowledge transfer difficult. Second, potential prior knowledge and physical information are typically ignored during heterogeneous transfer, hampering the utilization of domain-specific insights and leading to unstable mappings. Third, inappropriate information sharing among target and sources can easily lead to negative transfer. Traditional models fail to address these issues in a unified way. To overcome these limitations, this paper proposes a Double-Regularized Heterogeneous Gaussian Process framework (R²-HGP). Specifically, a trainable prior probability mapping model is first proposed to align the heterogeneous input domains. The resulting aligned inputs are treated as latent variables, upon which a multi-source transfer GP model is constructed and the entire structure is integrated into a novel conditional variational autoencoder (CVAE) based framework. Physical insights is further incorporated as a regularization term to ensure that the alignment results adhere to known physical knowledge. Next, within the multi-source transfer GP model, a sparsity penalty is imposed on the transfer coefficients, enabling the model to adaptively select the most informative source outputs and suppress negative transfer. Extensive simulations and real-world engineering case studies validate the effectiveness of our R²-HGP, demonstrating consistent superiority over state-of-the-art benchmarks across diverse evaluation metrics.

\end{abstract}

% % === KEYWORDS ====================================================================
% % =================================================================================

% \fi
\IEEEpeerreviewmaketitle

% ====================================================================
% ====================================================================
% ====================================================================

% === I. INTRODUCTION =============================================================
% =================================================================================
\section{Introduction}

\IEEEPARstart{G}{aussian}  process (GP), as a powerful nonparametric Bayesian framework, can flexibly models the function space through covariance functions and can naturally provides uncertainty quantification during prediction, achieving outstanding performance in numerous single-output tasks \cite{williams2006gaussian}, \cite{shi2011gaussian}, \cite{williams1995gaussian}. In engineering and scientific applications, there is often a need to predict multiple correlated outputs simultaneously, which has led to the development of multi-output Gaussian processes (MGP). MGP captures shared information among different outputs by incorporating cross-output correlations in the covariance structure and quantifies joint uncertainty. Typical MGP methods include the linear model of coregionalization (LMC) \cite{goulard1992linear}, \cite{alvarez2011computationally}, which represents each output as a linear combination of shared latent functions; the intrinsic coregionalization model (ICM), \cite{bonilla2007multi}, which assumes all outputs share the same base function and uses a coregionalization matrix for scaling coupling in the output dimension; the convolution process (CP) \cite{majumdar2007multivariate}, which generates non-stationary and deformable cross-output correlation structures through the convolution of white noise processes; and the Gaussian process regression network (GPRN) \cite{wilson2011gaussian}, which leverages a networked architecture to capture complex nonlinear and non-stationary correlations between outputs. These methods have been widely applied in fields such as computer vision, energy forecasting, bioinformatics, and engineering optimization, demonstrating strong modeling and generalization capabilities \cite{yue2024federated}, \cite{liu2018remarks}, \cite{gramacy2020surrogates}, \cite{liu2024gaussian}, \cite{ma2025probabilistic}.

However, in practical applications, the output data of some systems is often extremely scarce. The limited samples from a single system make it difficult for traditional multi-output Gaussian processes to accurately estimate the correlations and uncertainties between different outputs, significantly degrading their prediction performance \cite{cho2019hierarchical}. To overcome this bottleneck, researchers have introduced transfer learning schemes within the GP framework, transferring information from related source domains to enhance modeling capabilities in the target domain with limited data. The Kennedy-O’Hagan (KO) model \cite{KO} pioneered a GP-based co-kriging multi-fidelity modeling framework to integrate limited high-fidelity data with low-cost, low-fidelity simulation data. The recursive Co-Kriging model \cite{le2014recursive} extends the KO’s dual-fidelity approach to multi-level fidelity scenarios, where the lowest and second-lowest levels are linked using the KO model, and the prediction results serve as priors for higher levels. Kontar et al. \cite{kontar2018nonparametric} proposed a nonparametric transfer modeling framework based on multivariate Gaussian convolution processes for remaining useful life prediction. Wei et al. \cite{wei2022adaptive}, \cite{wei2022transfer} developed a series of transfer Gaussian process frameworks that explicitly model inter-domain relatedness through learnable transfer kernels, extending from single-source adaptive transfer to multi-source kernel learning. Yao et al. \cite{yao2025transfer} developed a transfer learning–based stochastic kriging model built upon a convolutional Gaussian process, which simultaneously models the within-process and between-process correlations to achieve individualized prediction. The common goal of the above methods is to leverage data from other related systems or domains to assist in target domain modeling, thereby effectively improving modeling performance in data-scarce scenarios. Similar to these methods, we focus on making predictions for one output which is denoted as the \textit{target}, by leveraging data from some related outputs which are denoted as \textit{sources} in this work.

Although the aforementioned transfer learning methods provide promising solutions to overcome the limitations posed by data scarcity, there are still some critical issues to be considered. First, existing methods often assume that the target and source share the same input domain, i.e., input domain homogeneity. However, in practical scenarios, the target and source may exhibit significant input domain heterogeneity, manifested as differences in feature dimensions, parameterization methods, or physical meanings \cite{pan2009survey}, \cite{zhou2014heterogeneous}. For example, in engineering design and simulation, due to variations in design complexity or fidelity, input domains are often inconsistent \cite{eppinger1995product}, \cite{sarkar2019multi}. Consider the design of aerodynamic blade as an example \cite{ghosh2022inverse}. Early design iterations primarily focus on optimizing efficiency. However, as the design evolves, additional features and details such as fixture features and locations, manufacturing constraints, and other functional considerations are introduced to address more nuanced requirements, changing the input parameter space in representation and complexity with each iteration. Take computational material science as another example. As the fidelity of structural simulations increases, the parameterization of material behaviors (e.g., elastic, plastic, composite materials) and analysis types (e.g., linear, nonlinear, static, dynamic) may also vary, further introducing heterogeneity among data sources \cite{nanoth2023static}. This heterogeneity often renders traditional transfer methods ineffective. Second, negative transfer easily arises in heterogeneous scenarios \cite{weiss2016survey}, \cite{marx2005transfer}, i.e., the jointly trained model may perform even worse than the one trained solely on the target data, which typically occurs when there is no shared information. The root cause of this phenomenon lies in the excessive inclusion of irrelevant or weakly related data during learning, a problem that becomes increasingly critical in the era of big data.

Several approaches have been proposed in the field of heterogeneous transfer learning and data fusion. Shi et al. \cite{shi2010transfer} proposed a low-dimensional projection framework for heterogeneous source and target domains, aiming to minimize the distributional discrepancy between the projected domains while preserving their respective data structures. Duan et al. \cite{duan2012learning} introduced a heterogeneous feature augmentation approach, which first maps source and target domain samples to a common subspace, then concatenates the mapped results with their original features and zero vectors to form augmented features. Wang and Mahadevan \cite{wang2011heterogeneous} jointly embeded the heterogeneous feature spaces of the source and target domains into a shared latent manifold , preserving each domain’s geometric structure while bringing samples of the same class closer to enable cross-domain classification. Input mapping calibration (IMC) \cite{tao2019input} fits a parameterized mapping from the target model’s input to that of the source model, aiming to match their outputs. IMC can derive physical insights from the mapping parameters. However, this method is limited to single-source scenarios and lacks the ability to adapt to multiple sources. Liu et al. \cite{liu2023learning} developed a heterogeneous multi-task Gaussian process, establishing a GP framework with multi-task capabilities for heterogeneous inputs. A recent contribution by Comlek et al. \cite{comlek2024heterogenous} combines IMC with latent variable Gaussian process (LVGP), providing a feasible solution for multi-source heterogeneous problems; however, this method trains the input mapping and data fusion steps separately, potentially leading to suboptimal performance. None of the above methods fully consider the uncertainty propagation in cross-domain alignment or the integration of known physical insights. Additionally, due to the risk for negative transfer, directly applying existing domain alignment methods to MGP may render transfer learning ineffective, as minimizing feature differences between negative sources and the target can exacerbate the severity of negative transfer. To the best of our knowledge, in the context of GP-based transfer learning, no work has simultaneously addressed heterogeneous domain alignment, end-to-end training, physical insights integration, and negative transfer mitigation in a unified framework.

To fill the gap, we propose a Double-Regularized Heterogeneous Transfer Gaussian Process framework (R²-HGP). Our approach focuses on aligning the target input space with each source input space, with the entire framework trained end-to-end via variational inference. Rather than treating all outputs equivalently as in many existing multi‐task learning, the proposed framework emphasizes directed knowledge transfer from every source output to the target output. The main contributions of this paper lie in the following three aspects: 
\begin{enumerate}[%
    label=\arabic*),           
    leftmargin=*,              
    labelsep=0.5em,                
    align=left   
]

  \item We propose the first general framework for heterogeneous transfer learning in the context of Gaussian Process (GP) multi-source transfer learning. Our approach uniquely handles the input heterogeneity in a unified manner, overcoming the constraints of traditional methods that rely on specific multi-fidelity hierarchies or shared feature assumptions.

  \item We develop a heterogeneous input domain alignment model for multi-source transfer learning, which is formulated to a novel conditional variational autoencoder (CVAE) framework. This model imposes a learnable prior distribution on the mapping, making it more flexible by accounting for domain alignment uncertainty.
  \item We design two novel regularization strategies. We introduce a physical-insight-based regularization term to ensure that the prior distribution of the mapping conforms to known physical knowledge; Moreover, we design a sparsity regularization term on the transfer coefficients, enabling the model to adaptively filter out detrimental sources, thereby proactively avoiding negative transfer.
  
\end{enumerate}

Our model achieves an end-to-end joint training framework for heterogeneous input domain alignment, physical insights integration, and negative transfer mitigation, effectively enhancing the stability and robustness of transfer learning. 

The remainder of this article is organized as follows. Section~\ref{sec:preliminaries}  introduces the general methodology for heterogeneous input space alignment and the formulated novel CVAE framework. Section~\ref{sec:meth} provides a detailed description of our double-regularized heterogeneous transfer Gaussian Process framework. Section~\ref{sec:experiments} presents numerical studies that demonstrate the superiority of the proposed method using both simulated and real data, and further conducts ablation experiments on the two regularization terms to validate their respective effects. Section~\ref{sec:conclu} concludes the paper.

\section{Preliminaries}\label{sec:preliminaries}
\subsection {Related Works}
Heterogeneous transfer learning has been extensively studied in deep learning, particularly those centered on classification and representation learning~\cite{day2017survey},~\cite{bao2023recent}. Most existing approaches are built upon deep neural network architectures, employing techniques such as feature representation mapping, cross-domain distribution alignment, shared subspace construction, or multimodal fusion to bridge the gap between source and target domains~\cite{zhou2019deep},~\cite{feuz2015transfer},~\cite{li2019heterogeneous}. These methods have achieved remarkable success in high-dimensional tasks such as natural language processing, computer vision, and multimodal learning~\cite{luo2017heterogeneous},~\cite{tian2025construction},~\cite{syu2025https}. However, much less attention has been paid to Gaussian process (GP) based heterogeneous transfer learning.

Within the context of GP-based transfer learning, only a few recent studies have addressed heterogeneity. Hebbal et al.~\cite{hebbal2021multi} introduced a non-parametric input mapping scheme to tackle heterogeneous issues in multi-fidelity transfer scenarios using a deep Gaussian process (DGP) model. Nevertheless, the training of this model relies entirely on an inaccurate nominal mapping, raising serious concerns about the reliability of its results. Wang et al.~\cite{wang2022regularized} proposed a domain adaptation through marginalization and expansion (DAME) method, which aligns heterogeneous input domains by marginalizing over the features shared between each source and the target, and expanding along the target-specific dimensions to generate pseudo-samples. On the one hand, this approach requires that the target and each source share at least one common feature in input domain, making it inapplicable to fully heterogeneous input domain. On the other hand, the generated pseudo-data may deviate from the true distribution. In addition, as the domain adaptation (DAME) and the subsequent GP modeling are carried out separately, the overall framework cannot achieve global optimality. A recent study~\cite{comlek2024heterogenous} combined IMC with latent variable Gaussian processes (LVGP), offering a feasible solution for heterogeneous data fusion with multi-source data. However, this approach not only lacks end-to-end integration, but the subsequent LVGP fusion is also unreliable: with limited data, the MLE-estimated latent variables easily overfit and create spurious source relationships, and the latent space further absorbs IMC mapping errors, causing the learned source differences to lose physical meaning and interpretability. GP-based transfer learning has also been applied in Bayesian optimization. Min et al.\cite{min2020generalizing} proposed a transfer Gaussian process model with a learned source feature transformation (Hetero-TGP), where a feature mapping is embedded in the transfer covariance function, mapping the source to a new representation that is aligned with that of the target task. Building on this, they developed a heterogeneous transfer Bayesian optimization (Hetero-TBO) algorithm to mitigate the cold-start problem and accelerate the convergence of black-box optimization. It is worth noting that heterogeneity can also arise in the output space~\cite{lee2024comprehensive},~\cite{xiao2024selective}. However, it is beyond the scope of our focus and will not be discussed here.

\subsection{Heterogeneous Input Mapping}\label{nominal}

In this subsection, we review some existing methods for heterogeneous input mapping. Let the target and source input–output pairs be denoted by $( \mathbf{x}^{tar}, y ^{tar})$ and $(\mathbf{x}^{S},y^{S})$, respectively.

One representative approach is to utilize a \emph{nominal mapping} $g_0(.):\mathbb{R}^{d_t}\mapsto\mathbb{R}^{d_s}$, where $d_t\leq d_s$ typically holds~\cite{tao2019input},~\cite{hebbal2019multi}. This nominal mapping expresses the presumed relationship between heterogeneous input spaces. In practice, it is typically specified based on expert knowledge or physical mechanisms, serving as a default correspondence between the source and target inputs. Since the nominal mappings are known in advance, a straightforward way is to map the task inputs to the source input domain for input alignment, and then perform the conventional MGP modeling in source domain, such as LMC~\cite{alvarez2011computationally}, ICM~\cite{bonilla2007multi}. A commonly adopted approach with a bias correction~\cite{li2016integrating} can also be used based on this nominal mapping:
$$
y^{tar}=f_{S}(g_0(\mathbf{x}^{tar}))+\gamma(\mathbf{x}^{tar}), 
\, \forall\,\mathbf{x}^{tar}\in\mathbb{R}^{d_{tar}},
$$
\noindent where $f_{S}(\cdot)$ denotes the model fitted to the source outputs, and $\gamma(\mathbf{x}^{tar})$ is the correction term.

In certain cases, the nominal mapping is trivial. For example, if both the source and target models describe the same physical system but differ only in modeling precision, the nominal mapping can be straightforward. Consider the thermal response of a metal plate: the source model may simply neglect secondary factors such as surface roughness or coating thickness, in which case its input variables can be directly regarded as a subset of those in the target model. Accordingly, the nominal mapping between their input domains can be directly defined by extracting the shared physical parameters. In general, however, the nominal mapping is problem-specific and often difficult to obtain.

Another representative method is input mapping calibration (IMC)~\cite{tao2019input}, which does not rely on the assumption of a known nominal mapping. IMC is a cross-domain calibration method, which is initially developed to reconcile the differing inputs of high-fidelity (target) and low-fidelity (source) models in multi-fidelity applications. Specifically, IMC introduces a parameterized mapping function $g\bigl(\mathbf{x}^{tar};\boldsymbol{\beta}\bigr)$ that projects the target input domain onto the source input domain. The mapping parameters $\boldsymbol{\beta}$ are then estimated by minimizing the discrepancy between the target outputs and the corresponding source outputs:

\begin{equation}
\label{Vanilla IMC}
    L(\boldsymbol{\beta})=\sum_{i=1}^N \big({y^{tar}}^{(i)}-y_{S}\big(g\bigl({\mathbf{x}^{tar}}^{(i)};\,\boldsymbol{\beta}\bigr)\big)\big)^2,
\end{equation}

\noindent where $N$ denotes the number of training samples. The parameterized mapping function $g\bigl(\mathbf{x}^{tar};\boldsymbol{\beta}\bigr)$ is typically chosen to be linear, i.e., $g(\mathbf{x}^{tar};\mathbf{A},\mathbf{b}) = \mathbf{A}\,\mathbf{x}^{tar} + \mathbf{b},$ where $A$ and $b$ denote the linear transformation matrix and bias vector, respectively.

IMC can provide physical insights~\cite{tao2019input}, as the estimated coefficients $\mathbf{A}$ and $\mathbf{b}$ can reveal how source inputs should be adjusted relative to target ones to compensate for the missing or simplified physics in the source model. Since the computation of $y_S$ is usually inexpensive, it is often replaced by a high-accuracy GP surrogate trained on a large number of source samples.

\subsection{Conditional Variational Autoencoder}
CVAE is an extension of the variational autoencoder (VAE) framework, designed to model complex data distributions conditioned on specific attributes or labels~\cite{sohn2015learning}. Unlike traditional VAEs, which learn a latent representation of data in an unsupervised manner, CVAEs incorporate conditional information, such as class labels or other contextual variables, into both the encoder and decoder networks. 

There are three types of variables in a CVAE: input variables $\mathbf{x}$, output variables $\mathbf{y}$, and latent variables $\mathbf{z}$. The conditional generative process of the model is as follows: for a given observed condition $\mathbf{x}$, $\mathbf{z}$ is drawn from the prior distribution $p_\theta(\mathbf{z}|\mathbf{x})$, and the output $\mathbf{y}$ is generated from the distribution $p_\theta(\mathbf{y}|\mathbf{x},\mathbf{z}).$ The latent variables $\mathbf{z}$ allow for modeling multiple modes in conditional distribution of output variables $\mathbf{y}$ given input $\mathbf{x}$. The prior of the latent variables $\mathbf{z}$ is modulated by the input $\mathbf{x}$ in formulation.
CVAE is trained to maximize the conditional log-likelihood. The objective function is often intractable, and the stochastic gradient variational Bayes (SGVB) framework is applied to train the model. The variational lower bound of the model can be derived as follows:
\begin{multline*}
    \log p_\theta(\mathbf{y}|\mathbf{x})\geq-KL\left(q_\phi(\mathbf{z}|\mathbf{x},\mathbf{y})\|p_\theta(\mathbf{z}|\mathbf{x})\right)\\+\mathbb{E}_{q_\phi(\mathbf{z}|\mathbf{x},\mathbf{y})}\Big[\log p_\theta(\mathbf{y}|\mathbf{x},\mathbf{z})\Big],
\end{multline*}

where $KL(\cdot\|\cdot)$ is the Kullback–Leibler divergence. CVAEs have been widely applied in domains such as image generation, where they can produce images with specific attributes (e.g., generating faces with specific expressions), and in natural language processing, for tasks like text generation conditioned on sentiment or topic. Their ability to incorporate conditional information makes them a powerful tool for modeling structured and controlled data generation.

\section{Proposed Methods}
\label{sec:meth}
In this section, we present the formalization of the proposed methods for heterogeneous transfer Gaussian process. The framework provides a flexible mechanism that simultaneously handles heterogeneous input domains, integrates physical insights, and mitigates negative transfer.
\subsection{Heterogeneous Domain Alignment}

The problem of transfer learning over heterogeneous input domains is defined as follows. We are given a set of source domains $\mathcal{S}=\{\mathcal{S}_i:1\leq i\leq N\}$ and one target domain $\mathcal{T}$, where each domain is defined as a pair of input and output: 
$\mathcal{S}_i=(\mathcal{X}_{\mathcal{S}_i},\mathcal{Y}_{\mathcal{S}_i})$ 
for a source domain and 
$\mathcal{T}=(\mathcal{X}_{\mathcal{T}},\mathcal{Y}_{\mathcal{T}})$ 
for the target domain. The collection of all domains is denoted as $\mathcal{D}=\{\mathcal{S}_1,\ldots,\mathcal{S}_N,\mathcal{T}\}$. For each domain, a finite dataset is sampled. Accordingly, we denote the data matrix and its corresponding label vector 
collected from each source domain $\mathcal{S}_i$ as 
$\mathbf{X}_i \in \mathbb{R}^{n_{\mathcal{S}_i}\times d_{\mathcal{S}_i}}$ 
and $\mathbf{y}_i \in \mathbb{R}^{n_{\mathcal{S}_i}\times 1}$, respectively, 
where $n_{\mathcal{S}_i}$ is the sample size and $d_{\mathcal{S}_i}$ is the feature dimension.
Likewise, for the target domain we denote 
$\mathbf{X}_T \in \mathbb{R}^{n_{\mathcal{T}}\times d_{\mathcal{T}}}$ 
as the input matrix and 
$\mathbf{y}_T \in \mathbb{R}^{n_{\mathcal{T}}\times 1}$ 
as the corresponding label vector. Source domains are assumed to have sufficient labeled data, whereas the target domain has only limited labels. More specifically, we have $n_{\mathcal{T}}\ll\sum_{i=1}^Nn_{\mathcal{S}_i}$. For the convenience, we define the concatenated input matrix and label vector of all the source and target domains as $\mathbf{X}= \{\mathbf{X}_1;\ldots;\mathbf{X}_{N};\mathbf{X}_{T}\}$ and $\mathbf{y}=(\mathbf{y}_1^T,\mathbf{y}_2^T,\ldots,\mathbf{y}_N^T,\mathbf{y}_T^T)^T$. Here $\mathbf{X}$ denote the collections of input matrices across all domains, used only as shorthand notation. Our objective is to utilize $\{\mathbf{X}_i, \mathbf{y}_i\}_{i=1}^N$ and $\{\mathbf{X}_T,\mathbf{y}_T\}$ to train a transfer model for the prediction of unseen target outputs $\mathbf{y}^*_T \in \mathbb{R}^{n^*_{\mathcal{T}}\times 1}$ 
at arbitrary new target inputs 
$\mathbf{X}^*_T \in \mathbb{R}^{n^*_{\mathcal{T}}\times d_{\mathcal{T}}}$.

The heterogeneity arises from the fact that the input domains of the sources, 
$\{\mathcal{X}_{\mathcal{S}_j}\}_{j=1}^N$, 
may differ from the input domain of the target, $\mathcal{X}_{\mathcal{T}}$, 
in their feature representations, 
i.e., there exists at least one source $j$ such that 
$\mathcal{X}_{\mathcal{S}_j}\neq \mathcal{X}_{\mathcal{T}}$. To address the heterogeneity issue, most existing methods employ deterministic linear domain mappings to project different heterogeneous domains into a low-dimensional common domain. Such deterministic projections, however, are prone to overfitting and may lead to information loss~\cite{zhang2021dense},~\cite{wani2025comprehensive}. For instance, a dataset that is linearly separable in its original input space may become nonlinear—or even non-separable—after undergoing a deterministic projection, thereby highlighting the potential risk of information loss caused by such rigid mappings~\cite{mcvay2022linear},~\cite{wang2021understanding}. 

To overcome the heterogeneous issue and the above limitation, we introduce a probabilistic prior on the input-domain alignment, which provides flexibility and capture uncertainty beyond a deterministic mapping.

Specifically, for a given target input $\mathbf{x}_T \in \mathcal{X}_{\mathcal{T}}$, 
we align it to each source input domain by introducing an aligned counterpart 
$\mathbf{x}_T^j$ in the input domain of source $j$, $\mathcal{X}_{\mathcal{S}_j}$. 
We then place a learnable prior distribution $p_{\theta_j}(\mathbf{x}_T^j \mid \mathbf{x}_T)$ over the aligned counterpart. Conceptually, we assume the existence of a fixed but unknown alignment function 
$\mathrm{Align}:\mathcal X_T \rightarrow \{\mathcal{X}_{\mathcal{S}_j}\}_{j=1}^N$ that maps each target input $\mathbf{x}_T$ to a set of latent aligned variables
 \begin{equation*}
    \mathbf{z}=\mathrm{Align}(\mathbf{x}_T) = 
    \left\{\mathbf{x}_T^{1},\ldots,\mathbf{x}_T^{N}\right\}
\end{equation*}The prior $p_{\theta}$ encodes our uncertainty about these aligned corresponding points. For tractability, we further assume that the conditional priors 
$\{p_{\theta_j}(\mathbf{x}_T^j \mid \mathbf{x}_T)\}_{j=1}^N$ 
are mutually independent. Consequently, the joint distribution of the aligned variables can be expressed as \begin{equation*}
    \mathbf{z}\mid \mathbf{x}_T \sim p_{\theta}(\mathbf{z}\mid \mathbf{x}_T) 
= \prod_{j=1}^N p_{\theta_j}(\mathbf{x}_T^j \mid \mathbf{x}_T)
\end{equation*}Here, each $\mathbf{x}_T^j$ is unobservable and is treated as a latent variable. The parameters of $p_\theta$ are jointly learned with the rest of the framework.

Let $\mathcal{I}^S=\{1,2,\ldots,N\}$ denotes the index set of sources, $\mathcal{I}=\{1,2,\ldots,N,T\}$ denotes the index set of sources and target. Based on the latent aligned values obtained through input-domain mapping, we construct the following heterogeneous multi-source transfer learning model:
\begin{equation}
\begin{aligned}
    y_{T}(\mathbf{x}_{T}) &= f_T(\mathbf{x}_{T}) + \epsilon_{T}(\mathbf{x}_{T}) 
    \\&= \sum_{j=1}^{N} \rho_j f_{j}(\mathbf{x}^{j}_{T}) + \delta_d(\mathbf{x}_{T})+\epsilon_{T}(\mathbf{x}_T), \\
    \delta_d(\mathbf{x}_T) &\sim \mathcal{GP}\left\{0, \mathcal{K}_d(\mathbf{x_T}, \mathbf{x_T^{\prime}})\right\}, \\
    f_j(\mathbf{x}_j) &\sim \mathcal{GP}\left\{0, \mathcal{K}_j(\mathbf{x}_j, \mathbf{x}_j^{\prime})\right\},\; j\in\mathcal{I}^S\\
    y_{S_j}(\mathbf{x}_j) &= f_j(\mathbf{x}_j) + \epsilon_j(\mathbf{x}_j), \; j\in\mathcal{I}^S\\
    \quad \delta_d &\perp f_j, \; j \in\mathcal{I}^S \\
    \quad f_{j'} &\perp f_j, \; j' \neq j \in \mathcal{I}^S\\
    \delta_d \perp &\, \epsilon_{T}\perp f_j, \; j \in\mathcal{I}^S,
\end{aligned}
\label{eq:transfer_model}
\end{equation}
where $y_{S_j}$ is the observed value of source domain $j$, $y_{T}$ is the observed value of target domain, $\delta_d(\mathbf{x_T})$ explains the potential differences between the target output and the weighted predictor using $N$ sources, and the symbol $\perp$ indicates that the corresponding Gaussian processes or the random variables are independent. $\epsilon_i(\mathbf{x})\sim\mathcal{N}(0,\sigma_i^2),i\in\mathcal{I}$ are i.i.d. Gaussian noises assigned to the $i$th output. Based on the assumption that $f(\mathbf{x})$ is independent of $\epsilon(\mathbf{x})$, the covariance between any two observations of the output $j\in\mathcal{I}^S$ can be decomposed as: $\textstyle \operatorname{Cov}\left(y_j\left(\mathbf{x}_j\right), y_{j}\left(\mathbf{x}_j^{\prime}\right)\right)=\operatorname{Cov}\left(f_j\left(\mathbf{x}_j\right), f_{j}\left(\mathbf{x}_j^{\prime}\right)\right)+\operatorname{Cov}\left(\epsilon_j\left(\mathbf{x}_j\right), \epsilon_{j}\left(\mathbf{x}_j^{\prime}\right)\right)=\mathcal{K}\left(\mathbf{x}_j,\mathbf{x}_j^{\prime}\right)+\sigma_j^2\tau\left(\mathbf{x}_j-\mathbf{x}_j^{\prime}\right)$, where $\tau\left(\mathbf{x}_j-\mathbf{x}_j^{\prime}\right)$ is equal to 1 if $\mathbf{x}_j=\mathbf{x}_j^{\prime}$, and 0 otherwise.

\begin{figure}[htbp]
\centering

\tikzset{
  dotnode/.style={circle, draw, minimum size=7mm, inner sep=1pt, align=center, font=\scriptsize},
  lossnode/.style={circle, draw, minimum size=7mm,  inner sep=1pt, align=center, font=\scriptsize},
  dashbox/.style={draw, dashed, inner sep=6pt, rounded corners=2pt},
}

\begin{tikzpicture}[
  >=Latex,
  node distance=10mm and 12mm,
  font=\small,
  every label/.style={font=\scriptsize, inner sep=1pt}
]

% ------- Left block: with prior prob. -------
\node[dotnode] (T) {$\mathcal{X}_{\mathcal{T}}$};

\node[dotnode, right=7mm of T, yshift=20mm] (S1) {$\mathcal{X}_{\mathcal{S}_1}$};
\node[dotnode, right=7mm of T, yshift=6mm ] (S2) {$\mathcal{X}_{\mathcal{S}_2}$};
\node[dotnode, right=7mm of T, yshift=-20mm] (SN) {$\mathcal{X}_{\mathcal{S}_N}$};

\node[font=\normalsize] (Svdots) at ($(S2)!0.5!(SN)$) {$\vdots$};

\node[font=\normalsize] (Lvdots) at ($(Svdots)+(-7mm,5mm)$) {$\vdots$};

% align arrows
\draw[->] (T) -- node[above,sloped] {\scriptsize align} (S1);
\draw[->] (T) -- node[above,sloped] {\scriptsize align} (S2);
\draw[->] (T) -- node[above,sloped] {\scriptsize align} (SN);

\node[dashbox, fit={(T) (S1) (SN) (Svdots) (Lvdots)}] (BOXL) {};
\node[anchor=south, font=\scriptsize, yshift=0.8mm] at (BOXL.north) {align with prior prob.\;\raisebox{-0.5ex}{\includegraphics[height=4.5mm]{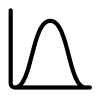}}
};

\node[dotnode, right=10mm of S1] (M1) {$f_{\mathcal{S}_1}$};
\node[dotnode, right=10mm of S2] (M2) {$f_{\mathcal{S}_2}$};
\node[dotnode, right=10mm of SN] (MN) {$f_{\mathcal{S}_N}$};

\node[font=\normalsize] (Mvdots) at ($(M2)!0.5!(MN)$) {$\vdots$};

% source input -> source model
\draw[->] (S1) -- (M1);
\draw[->] (S2) -- (M2);
\draw[->] (SN) -- (MN);

\coordinate (MidM) at ($(M1)!0.5!(MN)$);

\node[dotnode] (Y) at ($(MidM)+(12mm,0)$) {$\mathcal{Y}_{\mathcal{T}}$};

\draw[->] (M1) -- (Y);
\draw[->] (M2) -- (Y);
\draw[->] (MN) -- (Y);

\node[font=\normalsize] (Rvdots) at ($(Mvdots)+(7mm,5mm)$) {$\vdots$};

\node[dashbox, fit={(M1) (MN) (Mvdots) (Rvdots) (Y)}] (BOXR) {};
\node[anchor=south, font=\scriptsize, yshift=0.8mm] at (BOXR.north) {knowledge transfer};

% ------- Discrepancy -------
\node[lossnode, right=6mm of Y] (L) {$\delta(x)$};
\node[anchor=north] at ($(L.south)+(0,-1mm)$) {\scriptsize discrepancy};
\draw[<-] (Y) -- (L);

\end{tikzpicture}

\caption{The structure of our method with domain alignment (left) and knowledge transfer (right).}
\label{fig:structure-prior-kt}
\end{figure}
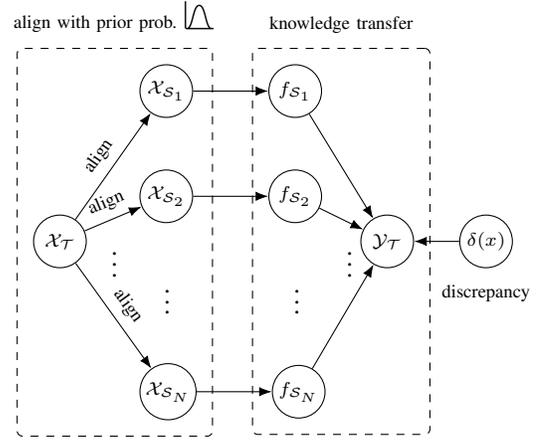

The structure of our model is illustrated in Fig.~\ref{fig:structure-prior-kt}. The data generation process can be described as follows. For all target inputs $\mathbf{X}_T$, there exist fixed but unobservable counterparts $\mathbf{X}_T^1$,$\mathbf{X}_T^2$,....,$\mathbf{X}_T^N$ in the respective source input domains, corresponding to $\mathbf{X}_T$. Without loss of generality, we assume that each sample in the target dataset has a distinct input, i.e., no duplicate inputs exist across target samples. Consequently, for each source input domain, there are $n_{\mathcal T}$ latent aligned inputs. These latent aligned inputs follow the priors defined previously. Equivalently, we may view $\mathbf{X}_T^1$,$\mathbf{X}_T^2$,....,$\mathbf{X}_T^N$ are drawn from the prior distribution 
$p_{\theta_1}({\mathbf{X}_T^1|\mathbf{X}_T)}, p_{\theta_2}({\mathbf{X}_T^2|\mathbf{X}_T)},...,p_{\theta_{N}}(\mathbf{X}_T^N|\mathbf{X}_T)$, where 
\begin{equation*}
\mathbf{X}_T^j=\left({\mathbf{x}_T^j}^{(1)},{\mathbf{x}_T^j}^{(2)},\dots, {\mathbf{x}_T^j}^{(n_{\mathcal{T}})}\right)^T.
\end{equation*}
 
\noindent We denote the collection of all such latent variables by $\mathbf{Z}=\left\{\mathbf{X}_T^1;\mathbf{X}_T^2;....;\mathbf{X}_T^N\right\}$, which represents the results of the domain feature mapping. Finally, the output $\mathbf{y}$ is generated from the distribution $P_{\theta_y}(\mathbf{y}|\mathbf{X},\mathbf{Z})$, which is defined in Eq.~\eqref{eq:transfer_model}. 

Our framework offers three advantages. First, we align target input domain separately to each source input domain. This preserves the integrity of source-domain features, without being discarded or compressed, thereby maximizing the utilization of physical meaning or structural information contained in the sources. Second, since each source’s contribution to the target is modeled independently, irrelevant sources can be down-weighted or excluded, effectively reducing the risk of negative transfer. Third, the model quantifies the uncertainty in the alignment. This probabilistic formulation alleviates the information loss that hard projections may incur, thereby improving robustness and enabling greater flexibility during domain alignment.

Based on Eq.~\eqref{eq:transfer_model}, we can define the covariance function of all outputs $\mathbf{y}$ as
\begin{equation}
\label{eq:cov-matrix}
\boldsymbol{C}_\mathbf{Z}(\mathbf{X},\mathbf{X})=\left(\begin{array}{cccc|c}
\boldsymbol{C}_{1,1} & \mathbf{0} & \cdots & \mathbf{0} & \boldsymbol{C}_{1, T} \\
\mathbf{0} & \boldsymbol{C}_{2,2} & \cdots & \mathbf{0} & \boldsymbol{C}_{2, T} \\
\vdots & \vdots & \ddots & \vdots & \vdots \\
\mathbf{0} & \mathbf{0} & \cdots & \boldsymbol{C}_{N, N} & \boldsymbol{C}_{N, T} \\
\hline \boldsymbol{C}_{1, T}^T & \boldsymbol{C}_{2, T}^T & \cdots & \boldsymbol{C}_{N, T}^T & \boldsymbol{C}_{T, T}
\end{array}\right)
\end{equation}
where each submatrix is as follows:
\begin{align*}
\boldsymbol{C}_{i,i} &= \boldsymbol{K}_i(\mathbf{X}_i,\mathbf{X}_i) + \sigma_i^2 I_{n_{S_i}},\quad \forall i\in \mathcal{I}^S \\
\boldsymbol{C}_{i,T} &= \rho_i \boldsymbol{K}_i(\mathbf{X}_i,\mathbf{X}_T^i), \quad \forall i\in \mathcal{I}^S \\
\boldsymbol{C}_{T,T} &= \sum_{j=1}^{N}\rho_j^2 \boldsymbol{K}_j(\mathbf{X}_T^j,\mathbf{X}_T^j) + \sigma_d^2 \boldsymbol{K}_d(\mathbf{X}_T,\mathbf{X}_T) + \sigma_T^2 I_{n_T} 
\end{align*} 

\noindent The proposed covariance function characterizes the cross-domain dependencies among all source and target outputs and serves as the covariance kernel of the proposed model. The validity of this function is outlined in Proposition~\ref{prop:pd}.

\begin{proposition}[Validity of the Covariance]\label{prop:pd}
The covariance function constructed under the proposed framework, denoted by $\mathbf{C}_{\mathbf{Z}}(\mathbf{X},\mathbf{X})$ in Eq.~\eqref{eq:cov-matrix}, is guaranteed to be strictly positive definite for any choice of real-valued weights $\{\rho_j\}$, thereby defining a valid Gaussian-process covariance kernel.
\end{proposition}

\noindent The proof is provided in Appendix A of the online supplementary material. Based on Eq.~\eqref{eq:transfer_model} and Eq.~\eqref{eq:cov-matrix}, the joint conditional distribution of all sources and the target is expressed as $\mathbf{y}\,|\,\mathbf{Z},\mathbf{X}\sim\mathcal{N}(\mathbf{0},\mathbf{C}_{\mathbf{Z}}(\mathbf{X},\mathbf{X}))$.

For notational convenience, let $\Theta = \{\theta_1, \theta_2, \ldots, \theta_N,\theta_y\}$ denote the collection of all the model parameters. The joint probability model of ($\mathbf{X}$,$\mathbf{y}$,$\mathbf{Z}$) is specified as $\;P_\Theta(\mathbf{X}, \mathbf{y}, \mathbf{Z})=P (\mathbf{X}) P_{\Theta} (\mathbf{Z}\mid\mathbf{X}) P_\Theta(\mathbf{y}\mid\mathbf{X}, \mathbf{Z})$. The prior term $P_\Theta(\mathbf{Z}\mid \mathbf{X})$ can take any suitable distribution, with a Gaussian distribution being the most common choice. The marginal conditional distribution for $\mathbf{y}$ is obtained via $P_\Theta(\mathbf{y} | \mathbf{X})=\int P_\Theta(\mathbf{Z} | \mathbf{X}) P_\Theta(\mathbf{y} | \mathbf{X}, \mathbf{Z}) d\mathbf{Z}$, and the exact posterior for $\mathbf{Z}$ is given by $P_\Theta(\mathbf{Z} |\mathbf{X}, \mathbf{y})=P_\Theta(\mathbf{y} \mid \mathbf{X}, \mathbf{Z}) P_\Theta(\mathbf{Z} \mid \mathbf{X})\,/\,P_\Theta(\mathbf{y} |\mathbf{X})$. The learning problem reduces to maximizing the conditional log-likelihood $\log P_\Theta(\mathbf{y}\mid \mathbf{X})$ and inferring the posterior of $\mathbf{Z}$, which is required for subsequent prediction. Since both $P_\Theta(\mathbf{y}\mid \mathbf{X})$ and $P_\Theta(\mathbf{Z}\mid \mathbf{X}, \mathbf{y})$ are analytically intractable, a variational posterior $Q_\Phi(\mathbf{Z}\mid \mathbf{X}, \mathbf{y})$ is introduced, typically parameterized as a Gaussian, to approximate $P_\Theta(\mathbf{Z}\mid \mathbf{X}, \mathbf{y})$. Notably, the resulting probabilistic structure naturally conforms to a novel 
\emph{Conditional Variational Autoencoder} (CVAE) paradigm, where the prior term $P_{\Theta} (\mathbf{Z} \mid \mathbf{X})$, the variational posterior $Q_\Phi(\mathbf{Z}\mid \mathbf{X}, \mathbf{y})$ and the conditional distribution $P_\Theta(\mathbf{y} \mid\mathbf{X}, \mathbf{Z})$ correspond to the conditional prior, encoder, and decoder, respectively. The target input $\mathbf{X}_T$ serves as contextual information that guides the latent alignment variables $Z$, and thus serves as the condition in the CVAE formulation. For clarity of exposition, we hereafter denote $P_{\Theta} (\mathbf{Z} \mid \mathbf{X})$ as the prior model, $Q_\Phi(\mathbf{Z}\mid \mathbf{X}, \mathbf{y})$ as the recognition model, and $P_\Theta(\mathbf{y} \mid \mathbf{X}, \mathbf{Z})$ as the observation model, which corresponds to the generative model in standard CVAE. It is worth emphasizing that, unlike conventional CVAE formulations which assume i.i.d. samples, our setting involves intrinsically correlated data and use a MGP as decoder. Consequently, the likelihood function and the following variational lower bound cannot be factorized across individual samples, requiring a more general treatment of the joint distribution.

We adopt the stochastic gradient variational Bayes (SGVB) framework \cite{kingma2013auto} to optimize the model. Specifically, the conditional likelihood $P_\Theta(\mathbf{y} | \mathbf{X})$  can be reformulated as follows:
\begin{equation}
    \begin{aligned}P_\Theta(\mathbf{y} | \mathbf{X})&=\int P_\Theta(\mathbf{y} | \mathbf{X}, \mathbf{Z})    
\frac{P_\Theta(\mathbf{Z} | \mathbf{X}) }{Q_\Phi(\mathbf{Z}\mid \mathbf{X}, \mathbf{y})}Q_\Phi(\mathbf{Z}\mid \mathbf{X}, \mathbf{y})d\mathbf{Z}\\&=\mathbb{E}_{\mathbf{Z}\sim Q_\Phi(\mathbf{Z}\mid \mathbf{X}, \mathbf{y})}\left\{P_\Theta(\mathbf{y} | \mathbf{X}, \mathbf{Z})    
\frac{P_\Theta(\mathbf{Z} | \mathbf{X}) }{Q_\Phi(\mathbf{Z}\mid \mathbf{X}, \mathbf{y})}\right\}.\end{aligned}
\label{eq:cond log}
\end{equation}

After taking the logarithm and applying Jensen's inequality, we obtain the variational lower bound of the conditional log-likelihood of all the observations:
\begin{equation}
\begin{aligned}
\log \big( P_\Theta(\mathbf{y} | \mathbf{X}) \big) 
&\geq \mathbb{E}_{Q_\Phi(\mathbf{Z}\mid \mathbf{X}, \mathbf{y})} \Bigg[
      \log \big( P_\Theta(\mathbf{y} | \mathbf{X}, \mathbf{Z}) \big) \\
&\quad + \log \frac{P_\Theta(\mathbf{Z} | \mathbf{X}) }{Q_\Phi(\mathbf{Z}\mid \mathbf{X}, \mathbf{y})}
   \Bigg]  \\
&= \mathbb{E}_{Q_\Phi(\mathbf{Z}\mid \mathbf{X}, \mathbf{y})} 
      \log \big( P_\Theta(\mathbf{y} | \mathbf{X}, \mathbf{Z})  \big) \\
&\quad - D_{KL}\!\Big(Q_\Phi(\mathbf{Z}\mid \mathbf{X}, \mathbf{y})\,\big\|\, 
   P_\Theta(\mathbf{Z} | \mathbf{X}) \Big). 
\end{aligned}
\label{eq:elbo}
\end{equation}
\noindent However, the first term in Eq.~\eqref{eq:elbo} only takes the sampling from recognition model $Q_\Phi(\mathbf{Z}\mid \mathbf{X}, \mathbf{y})$ during training. This may not be optimal during testing, since prediction for unseen target inputs $\mathbf{X}^*_T$ requires sampling from the prior $P_\Theta(\mathbf{Z}^* | \mathbf{X}^*_T)$ instead. In order to bridge the gap between training and testing phases, inspired by \cite{sohn2015learning},\cite{pandey2017variational}, this work trains the recognition model and the conditional prior model in a unified manner. This can be done by setting a hybrid lower bound. First, set $Q_\Phi(\mathbf{Z}\mid \mathbf{X}, \mathbf{y})=P_\Theta(\mathbf{Z} | \mathbf{X})$ in Eq.~\eqref{eq:cond log} and Eq.~\eqref{eq:elbo}, resulting in an alternative lower bound: 
 $$\log \big( P_\Theta(\mathbf{y} | \mathbf{X}) \big)\geq\mathbb{E}_{P_\Theta(\mathbf{Z} | \mathbf{X}) } 
      \log \big[ P_\Theta(\mathbf{y} | \mathbf{X}, \mathbf{Z})\big]$$
\noindent Next, the lower bounds of two equations are combined to obtain a hybrid lower bound as Eq.~\eqref{eq:doublerec}.

\begin{equation}
\begin{split}
\log \big( P_\Theta(\mathbf{y} | \mathbf{X}) \big) 
&\geq \mu \Big\{ - D_{KL}\!\Big(Q_\Phi(\mathbf{Z}\mid \mathbf{X}, \mathbf{y})\,\big\|\, 
   P_\Theta(\mathbf{Z} | \mathbf{X}) \Big)\Big\} \\
&\quad + \mu \,\mathbb{E}_{Q_\Phi(\mathbf{Z}\mid \mathbf{X}, \mathbf{y})} 
      \log \big[ P_\Theta(\mathbf{y} | \mathbf{X}, \mathbf{Z})  \big] \\
&\quad + (1-\mu)\,\mathbb{E}_{P_\Theta(\mathbf{Z} | \mathbf{X}) } 
      \log \big[ P_\Theta(\mathbf{y} | \mathbf{X}, \mathbf{Z})  \big] .
\end{split}
\label{eq:doublerec}
\end{equation}

\noindent where the reconstruction is performed using samples from both $Q_\Phi(\mathbf{Z}\mid \mathbf{X}, \mathbf{y})$ and $P_\Theta(\mathbf{Z} | \mathbf{X})$. The mixing weight $0 \leq \mu \leq 1$ balances the two objectives during training stage. Its value is typically chosen empirically depending on the particular scenario.

To train the CVAE-based framework, we construct the objective function based on the combination of terms in Eq.~\eqref{eq:doublerec} as:
\begin{equation}
\label{eq:loss CVAE}
\mathcal{L}_{CVAE} = \mathcal{L}_{KL} + \mathcal{L}_{rec}.
\end{equation}

\noindent The distribution regularization term $\mathcal{L}_{KL}$  aims to regularizes consistency between the prior of the aligned latent variables and their variational posterior encoded
from the observed input–output pairs:
\begin{equation}
\label{eq:KL}
\mathcal{L}_{KL}=\mu \Big\{ - D_{KL}\!\Big(Q_\Phi(\mathbf{Z}\mid \mathbf{X}, \mathbf{y})\,\big\|\, 
   P_\Theta(\mathbf{Z} | \mathbf{X}) \Big)\Big\}.
\end{equation}
The reconstruction terms $\mathcal{L}_{rec}$ denotes the likelihood components in Eq.~\eqref{eq:doublerec}, encouraging accurate reconstructions of the outputs $\mathbf{y}$. Particularly, the reconstruction terms can be approximated by drawing latent variables samples 
$\mathbf{Z}^{(l)} \ (l=1,\ldots,L)$, $\mathbf{Z}^{(m)} \ (m=1,\ldots,M)$  from the recognition distribution $Q_\Phi(\mathbf{Z}\mid\mathbf{X},\mathbf{y})$ and the prior distribution $P_\Theta(\mathbf{Z}\mid\mathbf{X})$ respectively. Accordingly, $\mathcal{L}_{rec}$ is defined as the empirical objective of the reconstruction likelihood, given by:\begin{equation}
\begin{aligned}
{\mathcal{L}}_{rec} 
&= \mu\, \frac{1}{L}\sum_{l=1}^L 
    \log \big[ P_\Theta(\mathbf{y} \mid \mathbf{X}, \mathbf{Z}^{(l)}) \big] \\
&\quad +\,(1-\mu)\,\frac{1}{M}\sum_{m=1}^M 
    \log \big[ P_\Theta(\mathbf{y} \mid \mathbf{X}, \mathbf{Z}^{(m)}) \big].
\end{aligned}
\label{eq:loss rec}
\end{equation}

\noindent Note that the recognition distribution $Q_\Phi(\mathbf{Z}\mid\mathbf{X},\mathbf{y})$ and the prior distribution $P_\Theta(\mathbf{Z}\mid\mathbf{X})$ are reparameterized with deterministic, 
differentiable functions $h_{\Phi}^Q(.)$ and $h_{\Theta}^P(.)$, whose arguments include the noise variable $\epsilon$. Specifically, $\mathbf{Z}^{(l)} =h_{\Phi}^Q(\boldsymbol{\epsilon},\mathbf{X},\mathbf{y}),\boldsymbol{\epsilon}\sim q_{\boldsymbol{\epsilon}}(\boldsymbol{\epsilon})$ and $\mathbf{Z}^{(m)} =h_{\Theta}^P(\boldsymbol{\epsilon},\mathbf{X}),\boldsymbol{\epsilon}\sim q_{\boldsymbol{\epsilon}}^{\prime}(\boldsymbol{\epsilon})$, which implies that $\mathbf{Z}^{(l)}\sim Q_{\Phi}(\mathbf{Z}|\mathbf{X},\mathbf{y})$ and $\mathbf{Z}^{(m)}\sim P_{\Theta }(\mathbf{Z}|\mathbf{X})$. This reparameterization trick allows error backpropagation efficiently, which is essential in training.
\subsection{Physical-Insight Regularization}
During the alignment of heterogeneous input domains, relying exclusively on data-driven learning approaches may introduce excessive flexibility, which could result in outcomes that deviate from fundamental physical insights. Conversely, known physical principles or domain knowledge can offer valuable insights into uncovering relationships between input domains. Incorporating this knowledge during training can enhance both the validity and robustness of the alignment. A specific real-world example\cite{tao2019input} is the aerodynamic simulation of an aircraft wing: in the high-fidelity (HF) vortex ring method (VRM), the wing is accurately modeled as a three-dimensional structure with thickness; in contrast, in the low-fidelity (LF) vortex lattice method (VLM), the thickness is neglected, and the wing is approximated as a two-dimensional cambered surface. Thus, the input for the HF model includes airfoil thickness, while the LF model does not. In this case, the physically informed relationship between the heterogeneous input domains is established by projecting the three-dimensional wing geometry of the HF model onto its mean camber surface, an approach that is theoretically justified.

The physical insight can be represented by a functional form, denoted as $r_0(\cdot)$, hereinafter referred to as the reference mapping. To incorporate this reference mapping, we introduce a physical-insight regularization term as follows:
\begin{equation}
\label{eq:phyR}
\begin{aligned}
\mathcal{L}_{PhyR}=\mathbb{R}_\lambda(\Theta)&=-\lambda\sum_{j=1}^{N}\Omega(\theta_j)\\&=-\lambda\sum_{j=1}^{N}\left\|r_j\left(\mathbf{X}_T,\theta_j\right)-r_{0j}\left(\mathbf{X}_T\right)\right \|_F
\end{aligned}
\end{equation}
where $r_j\left(\mathbf{X}_T,\theta_j\right),\,r_{0j}\left(\mathbf{X}_T\right)\in \mathbb{R}^{n_{\mathcal{T}} \times d_{\mathcal{T}} }$, with $r_j\!\left(\mathbf{X}_T,\theta_j\right)=\mathbb{E}_{\mathbf{X}_T^j}\,p_{\theta_j}(\mathbf{X}_T^j\mid\mathbf{X}_T)$ 
denoting the expectation under the prior model of source $j$, 
and $r_{0j}\!\left(\mathbf{X}_T\right)$ representing the physical-insight-based 
reference aligned value of $\mathbf{X}_T$ for source $j$. 
Here, $\|\cdot\|_F$ denotes the Frobenius norm.

Our physical-insight-based reference mapping $r_{0j}(\cdot)$, $j\in \mathcal{I}^S$ can be obtained in two ways. When explicit physical mechanisms or domain knowledge are available, a nominal mapping can be specified by such prior knowledge, as discussed above and in Section~\ref{nominal}, and therefore selected as the reference mapping. In contrast, when such knowledge is absent, IMC can be utilized to extract physical insights~\cite{tao2019input,comlek2024heterogenous}. By revealing how the target input domain should be adjusted to compensate for the missing physical effects in the source domain, it effectively achieves a form of “rediscovery” of physical insights. In this way, IMC provides new and interpretable physical insights even when explicit physical knowledge or expert experience is unavailable. As demonstrated in Section~\ref{sec:Ablation Studies}, this strategy successfully compensates for cases where subject-matter expertise is lacking.

\subsection{Source-Selection Regularization}
Negative transfer is more likely to occur under heterogeneous conditions~\cite{zhang2022survey}. To ensure the robustness of the transfer learning process, we introduce a source selection mechanism, which suppresses irrelevant or noisy sources, thereby highlighting informative ones and improving the overall generalization performance. It is worth noting that if the weight parameter $\rho_i$ is penalized to zero, the potential negative transfer between $f_T$ and $f_i$ can be completely avoided. To enforce sparsity by driving the transfer
coefficients that connect the target to uncorrelated sources
toward zero, we propose the Source-Selection Regularization:\begin{equation}
\label{eq:ssr}
\mathcal{L}_{SSR}=\mathbb{R}_\gamma(\boldsymbol{\rho})
\end{equation}

\noindent where$\boldsymbol{\rho}=(\rho_1,\rho_2,\ldots,\rho_N)$, and $\mathbb{R}_\gamma(\boldsymbol{\rho})$ denotes a non-positive penalty function. Common choices for the penalty include the $L_1$ norm, $\mathbb{R}_\gamma(\boldsymbol{\rho})=-\gamma\|\boldsymbol{\rho}\|_{1}$, and the smoothly clipped absolute deviation (SCAD) function~\cite{fan2001variable}.

Finally, we obtain the overall objective function by combining the terms in Eqs.~\eqref{eq:loss CVAE},~\eqref{eq:phyR} and~~\eqref{eq:ssr}:
\begin{equation}
\label{eq:overal objective}
\begin{aligned}
\mathcal{L}\left( \Theta ,\Phi ;\mathbf{X},\mathbf{y}\right)&=\mathcal{L}_{CVAE}+\mathcal{L}_{PhyR}+\mathcal{L}_{SSR}\\
    &= \mu\, \Big\{ - D_{KL}\!\Big(Q_\Phi(\mathbf{Z}\mid \mathbf{X}, \mathbf{y})\,\big\|\, 
   P_\Theta(\mathbf{Z} | \mathbf{X}) \Big)\Big\} \\
&\quad+ \mu\, \frac{1}{L}\sum_{l=1}^L 
    \log \big[ P_\Theta(\mathbf{y} \mid \mathbf{X}, \mathbf{Z}^{(l)}) \big] \\
&\quad +\,(1-\mu)\,\frac{1}{M}\sum_{m=1}^M 
    \log \big[ P_\Theta(\mathbf{y} \mid \mathbf{X}, \mathbf{Z}^{(m)}) \big]\\
    &\quad -\lambda\sum_{j=1}^{N}\left\|r_j\left(\mathbf{X}_T,\theta_j\right)-r_{0j}\left(\mathbf{X}_T\right)\right \|_F\\
    &\quad+ \mathbb{R}_\gamma(\boldsymbol{\rho})\\
\end{aligned}
\end{equation}

\begin{figure*}[ht!]
\centering
\includegraphics[width=\textwidth]{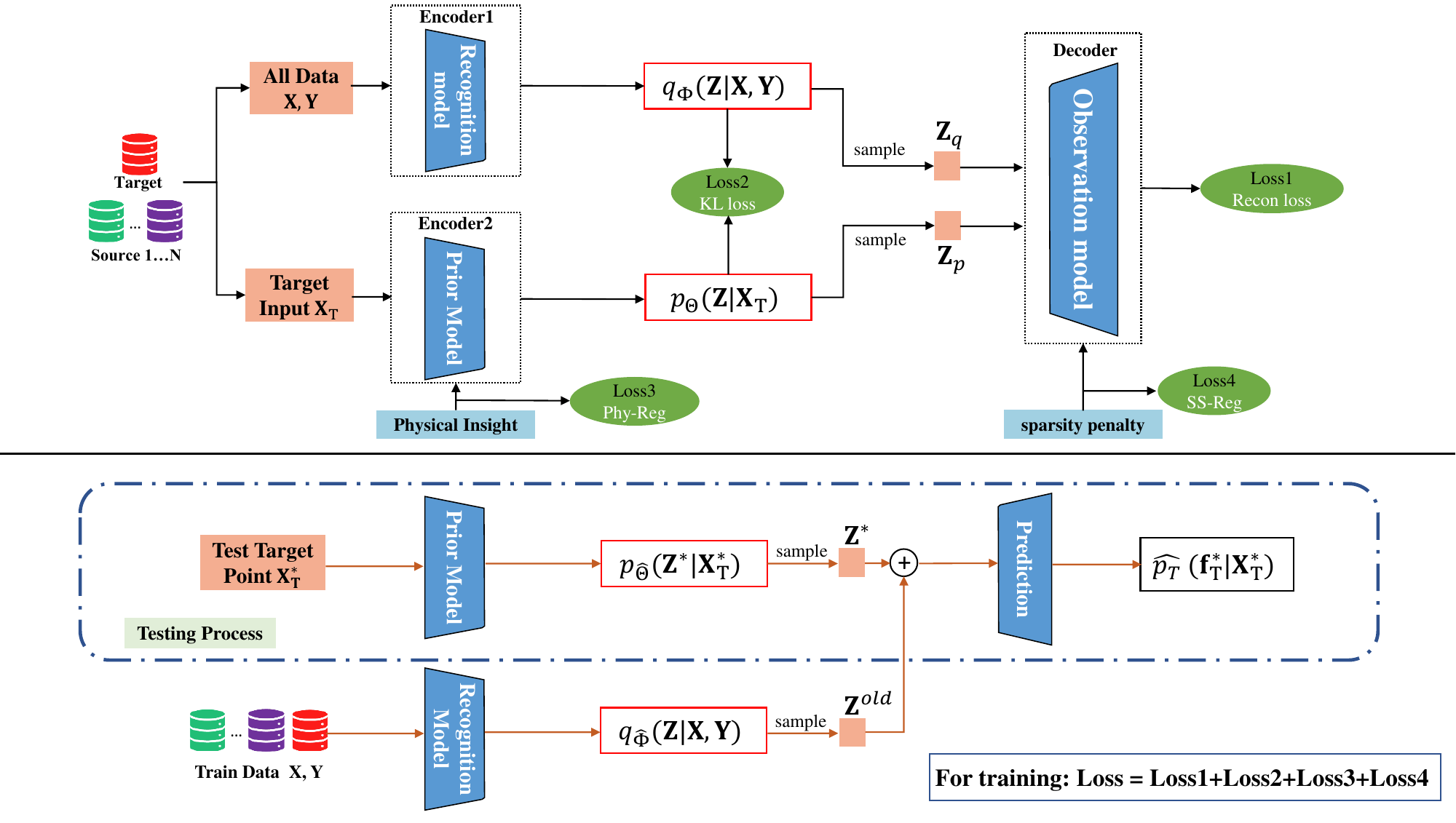}
\caption{Overview of our proposed framework, with the upper part showing the training process and the lower part showing the testing process.}
\label{fig:overview}
\end{figure*}

\noindent Our goal is to maximize $\mathcal{L}\left( \Theta ,\Phi ;\mathbf{X},\mathbf{y}\right)$ with respect to the parameters $\Theta$ and $\Phi$. We refer to the proposed approach as R$^2$-HGP (Double-Regularized Heterogeneous Transfer Gaussian Process). Fig.~\ref{fig:overview} depicts the overview of our proposed method architecture.

\subsection{Prediction}

Once the model parameters have been learned, we perform out-of-sample prediction for unseen target inputs. Specifically, we employ the inferred posterior distribution of latent variables, together with the estimated parameters, to construct the predictive posterior distribution of the target outputs $\mathbf{f}_T^* \in \mathbb{R}^{n^*_{\mathcal{T}}\times 1}$ given new target inputs $\mathbf{X}^*_T \in \mathbb{R}^{n^*_{\mathcal{T}}\times d_{\mathcal{T}}}$ and the observed training data.

Let the estimated parameters be denoted by $\widehat{\Theta}$ and $\widehat{\Phi}$. Conditional on the observed data $(\mathbf{X},\mathbf{y})$, the corresponding latent variables $\mathbf{Z}$, the estimated parameters $(\widehat{\Theta}, \widehat{\Phi})$, and the domain-aligned latent variables $\mathbf{Z}^*=\big\{{\mathbf{X}_{T}^1}^\ast ,{\mathbf{X}_{T}^2}^\ast ,\ldots {\mathbf{X}_{T}^{N}}^\ast \big\}$ for the new target inputs $\mathbf{X}^*_T$, the predictive posterior distribution of the target outputs $\mathbf{f}_T(\mathbf{X}^*_T)$ follows a Gaussian distribution:\begin{equation}
\label{eq:predictive_f_T}
\mathbf{f}_T(\mathbf{X}^*_T)|\mathbf{X},\mathbf{y},\mathbf{Z},\mathbf{Z}^*\sim\mathcal{N}(\boldsymbol\mu_T(\mathbf{X}^*_T),\mathbf{V}_T(\mathbf{X}^*_T))
\end{equation}
where,\begin{equation}
\label{eq:predictive_mean var}
    \begin{aligned}
        \boldsymbol\mu_T(\mathbf{X}^*_T)&=cov_{\mathbf{Z}^*}^T(\mathbf{X},\mathbf{X}^*_T)\boldsymbol{C}_\mathbf Z (\mathbf X,\mathbf X)^{-1}\mathbf{y},\\\mathbf{V}_T(\mathbf{X}^*_T)&=cov_{\mathbf{Z}^*}(\mathbf{X}^*_T,\mathbf{X}^*_T)\\&-cov_{\mathbf{Z}^*}^T(\mathbf{X},\mathbf{X}^*_T)\boldsymbol{C}_\mathbf Z (\mathbf X,\mathbf X)^{-1}cov_{\mathbf{Z}^*}(\mathbf{X},\mathbf{X}^*_T)
    \end{aligned}
\end{equation}

\noindent Due to the stochasticity of $\mathbf{Z}$ and $\mathbf{Z}^*$, we further marginalize over these latent variables to obtain the approximate predictive posterior of $\mathbf{f}_T^*$:\begin{multline}
\label{prediction}
    p\bigl(\mathbf{f}_T^* \mid \mathbf{X}_T^*, \mathbf{X}, \mathbf{y}\bigr)
\approx
\iint
P_{\widehat{\Theta}}\bigl(\mathbf{Z}^*\mid \mathbf{X}_T^*\bigr)\,Q_{\widehat{\Phi}}(\mathbf{Z}\mid \mathbf{X}, \mathbf{y})\\
p_{\widehat{\Theta}}\bigl(\mathbf{f}_T^*\mid \mathbf{X}_T^*,\mathbf{Z}^*,\mathbf{Z},\mathbf{X},\mathbf{y}\bigr)
\,\mathrm{d}\mathbf{Z}\,\mathrm{d}\mathbf{Z}^*.
\end{multline}

In Eq.~\eqref{prediction}, the intractable exact posterior over $\mathbf{Z}$ is replaced with its variational approximation $Q_{\widehat{\Phi}}(\mathbf{Z}\mid \mathbf{X}, \mathbf{y})$(see Appendix B for details). The conditional term $p_{\widehat{\Theta}}\bigl(\mathbf{f}_T^* \mid \mathbf{X}_T^*,\mathbf{Z}^{*},\mathbf{Z},\mathbf{X},\mathbf{y}\bigr)$ is computed according to Eq.~\eqref{eq:predictive_f_T}. Based on this procedure, 
the overall prediction workflow is illustrated in Fig.~\ref{fig:overview}.

\subsection{Model Implementation}

This section presents the implementation details of our model, including the choice of kernel functions, the design of the prior and recognition models, and the optimization strategies adopted.

For each source \(j\) and target sample \(i\), the posterior of $\mathbf{x}_T^{j,(i)}$ should, in principle, depend on all observed data and be mutually coupled. Following~\cite{dai2015variational}, we make a tractable approximation by fully factorizing the variational distribution. Accordingly, the recognition model \(Q_\Phi(\mathbf Z\mid \mathbf X,\mathbf Y)\) is modeled as a fully factorized Gaussian over each source–sample pair \((j,i)\). The mean and variance functions in both the prior model $P_\Theta(\mathbf Z\mid \mathbf X)$ and the recognition model $Q_\Phi(\mathbf Z\mid \mathbf X,\mathbf y)$ are parameterized as learnable mappings. With these settings, efficient sampling and closed-form computation of the KL divergence can be achieved for both models (see Appendix C for details).

For the kernel choice, we employ the widely used squared exponential (SE) kernel for both $\mathcal{K}_d(\cdot, \cdot)$ and $\mathcal{K}_j(\cdot, \cdot),, j\in\mathcal{I}^S$. The SE kernel provides a smooth and infinitely differentiable prior, which ensures continuity and effectively captures correlations among inputs:
\begin{equation}
\label{eq:kernel}
    \mathcal{K}(\mathbf{x}, \mathbf{x}') = \alpha^2 \exp\!\left(-\tfrac{1}{2} (\mathbf{x} - \mathbf{x}')^{T} \mathbf{\Lambda}^{-1} (\mathbf{x} - \mathbf{x}') \right),
\end{equation}

\noindent where $\alpha$ and $\mathbf{\Lambda}$ are the amplitude and length-scale hyper-parameters, respectively.

Finally, we optimize the overall objective function in Eq.~\eqref{eq:overal objective} using the Adam optimizer~\cite{kingma2014adam}. To promote sparsity and suppress uninformative sources, We impose $L_1$ penalty on the vector of transfer coefficients $\rho$. For hyperparameter tuning, the parameters $(\lambda,\gamma)$ are selected via cross‐validation. Specifically, we search over candidate values and choose the combination that yields the lowest validation RMSE, evaluated under either leave‐one‐out or 5‐fold cross‐validation. As for the hyperparameter $\mu$, its value is typically chosen empirically; we set \(\mu = 0.7\) for most cases in our experiments.

The implementation of the Double-Regularized Heterogeneous Gaussian Process (R$^2$-HGP) is summarized in Algorithm~\ref{alg:r2hgp}, while a detailed flowchart illustrating the full workflow is included in the Appendix C.

\begin{algorithm}[t]
\caption{Procedure of the Double-Regularized Heterogeneous Transfer Gaussian Process (R$^2$-HGP) Modeling}
\label{alg:r2hgp}
\begin{algorithmic}[1]
\Require Multi-source datasets $\{(\mathbf{X}_i,\mathbf{y}_i)\}_{i=1}^N$, target dataset $(\mathbf{X}_T,\mathbf{y}_T)$
\Ensure Estimated parameters $(\widehat{\Theta},\widehat{\Phi})$ and predictions $\{\mu_T(\mathbf{x}_T^\ast),\sigma_T^2(\mathbf{x}_T^\ast)\}$

\State \textbf{Initialization:} 
Initialize $P_\Theta(\mathbf{Z}\mid \mathbf{X})$, $Q_\Phi(\mathbf{Z}\mid \mathbf{X},\mathbf{y})$, and $P_\Theta(\mathbf{y}\mid \mathbf{X},\mathbf{Z})$ with initial parameters.

\For{each source–target pair $j = 1, \dots, N$}
  \If{explicit physical insight is available}
     \State Set reference mapping $r_{0j}(\mathbf{X}_T)$
  \Else
     \State Apply IMC to obtain reference mapping $r_{0j}^{\text{IMC}}(\mathbf{X}_T)$ for physical insights extraction
  \EndIf
\EndFor

\State \textbf{Aggregation:} Form $\{r_{0j}(\mathbf{X}_T)\}$ by taking $r_{0j}$ when available, and $r_{0j}^{\text{IMC}}$ otherwise.

\State \textbf{Optimization:} Maximize Eq.~\eqref{eq:overal objective} using the Adam optimizer to obtain $(\widehat{\Theta},\widehat{\Phi})$.

\State \textbf{Prediction:} For a set of new inputs $\mathbf{X}_T^\ast \subseteq \mathcal{X}_{\mathcal{T}}$:
\State \hspace{1em} Sample $\mathbf{Z}^\ast \sim P_{\widehat{\Theta}}$ and $\mathbf{Z} \sim Q_{\widehat{\Phi}}$
\State \hspace{1em} Compute predictive mean and covariance $\mu_T(\mathbf{X}_T^\ast)$, $\sigma_T^2(\mathbf{X}_T^\ast)$ via Eq.~\eqref{eq:predictive_f_T}--\eqref{eq:predictive_mean var}

\State \Return $(\widehat{\Theta}, \widehat{\Phi})$ and predictions
\end{algorithmic}
\end{algorithm}

\section{Case Studies}
\label{sec:experiments}

We next evaluate the proposed double–regularized heterogeneous multi–source transfer model, R$^2$-HGP, using three simulation case studies and three real-world engineering problems. Section~\ref{sec:General Settings} presents the experimental setup and baseline methods. Section~\ref{sec:Simulation Case} demonstrates the advantages of R$^2$-HGP in incorporating physical insights and mitigating negative transfer in simulation tasks. Section~\ref{sec:Real Case} reports results on real-world engineering applications, and Section~\ref{sec:Ablation Studies} provides an ablation analysis of the two regularization terms.
\subsection {General Settings}\label{sec:General Settings}
For comparison, we evaluate our approach against five benchmark methods:
\begin{enumerate}
  \item \textbf{IMC}~\cite{tao2019input}: Input Mapping Calibration, which linearly maps the target input to the domain of a single source. Since IMC can only handle one source at a time, we select in each case either the source with the largest amount of data or the one most similar to the target.

  \item \textbf{MF-DGP-EM}~\cite{hebbal2021multi}: A multi-fidelity Deep Gaussian Process with embedding mapping, which requires the target and source domains to be distinguished by fidelity levels that obey a strict hierarchical ordering. It further assumes the existence of nominal mapped values; consequently, this method is only applicable in a limited range of scenarios.

  \item \textbf{MGCP-R}~\cite{wang2022regularized}: 
  The Domain Adaptation through Marginalization and Expansion (DAME) method, which generates pseudo-samples along target-specific features, and is further incorporated into a regularized MGCP framework. This method requires that each source share at least one common input feature with the target, which thus limits its wide applications.
        
  \item \textbf{SA-LVGP}~\cite{comlek2024heterogenous}: A two-stage method denoted as SA-LVGP. In the first stage, the IMC algorithm projects all heterogeneous input domains into a unified reference domain. In the second stage, an LVGP-based multi-source data fusion model is trained on this unified space to construct a single source-aware surrogate model. 
  
  \item \textbf{TGP}: A single-output Gaussian Process model trained solely on target-domain data, denoted as TGP.
  
\end{enumerate}

We evaluate the above methods on the test set $\{\mathbf x_T^{*(i)},\,y_T(\mathbf x_T^{*(i)})\}_{i=1}^{n_{\text{test}}}$. To quantify predictive performance, three complementary criteria are adopted: Root Mean Squared Error (\textbf{RMSE})~\cite{chai2014root}, Coefficient of Determination (\textbf{$\boldsymbol{R}^2$})~\cite{chicco2021coefficient}, Mean Negative Log-Likelihood (\textbf{MNLL})~\cite{lind2024uncertainty}. \textbf{RMSE} quantifies the average deviation between predicted and observed values; smaller values indicate higher predictive accuracy. \textbf{$\boldsymbol{R}^2$} measures the proportion of variance in the data explained by the model; values closer to~1 imply a better fit. \textbf{MNLL} evaluates how well the predictive distribution captures both accuracy and uncertainty. The computation is based on the log-likelihood of the entire output vector under the joint predictive distribution, thereby explicitly incorporating the covariance structure \({\mathbf V}_T\). This allows MNLL to reflect the quality of correlated uncertainty quantification across test points. Lower values indicate better predictive performance.

Gaussian kernel in Eq.~\eqref{eq:kernel} is used for all methods. All trainable parameters in our framework are optimized using the Adam optimizer ~\cite{kingma2014adam} with default momentum parameters and an initial learning rate of $2 \times 10^{-3}$. All kernels’ initial length-scales and amplitudes are set to 1, the transfer coefficients $\rho$ are also initialized to 1,  noise parameter $\sigma$ for all outputs are initialized with random values in $[0,1]$. More details can be found in Appendix D.

\subsection {Simulation Cases}\label{sec:Simulation Case}

In all simulation studies, we assumed that physical insights about input domain mappings were available and were derived from the functional forms of the simulation models. Notably, the assumed physical insights contain no prior knowledge about which sources may induce negative transfer; instead, the model adaptively learns to down-weight or exclude negative sources.

To evaluate the robustness of our method, we repeated the sampling of training data 30 times using Latin Hypercube Sampling (LHS). For each repetition, the model was retrained and evaluated, and the mean $\pm$ one standard deviation of all metrics were reported.

\subsubsection {Simulation Case 1}

The first simulation case adapts the well known Park multi-fidelity problem \cite{park1991tuning} to the heterogeneous setting. Park function is used widely in multi-fidelity research as a benchmark problem. This case illustrates the situations where some variables are neglected in the low-fidelity (LF) model for simplicity. Our goal is to verify whether the proposed model can effectively leverage information from reduced or misspecified LF models to boost predictions when the LF variables are a subset of high-fidelity (HF) variables. Accordingly, one target response and three auxiliary sources ($\mathcal S_1,\mathcal S_2,\mathcal S_3$) are constructed. The target input domain is a four-dimensional unit hypercube, and each source domain corresponds to its respective input subspace:\begin{align*}
&\mathcal{X}_{\mathcal T}:\;\mathbf x_T = (x_1,x_2,x_3,x_4)\in[0,2]^4,\\
&\mathcal{X}_{\mathcal S_1}:\;\mathbf x_{S_1}=(x_1,x_2)\in[0,2]^2,\\
&\mathcal{X}_{\mathcal S_2}:\;\mathbf x_{S_2}=(x_1,x_2)\in[0,2]^2,\\
&\mathcal{X}_{\mathcal S_3}:\;\mathbf x_{S_3}=(x_1,x_2,x_3)\in[0,2]^3.
\end{align*}
\noindent The target output is:
\begin{multline*}
  f_T(\mathbf x_T) =
    \frac{x_1}{2}\Bigl(\sqrt{1+(x_2+x_3^{2})\tfrac{x_4}{x_1^{2}}}-1\Bigr)\\
    +\;(x_1+3x_4)\exp\,\bigl(1+\sin x_3\bigr),
\end{multline*}
\noindent with three auxiliary sources :
% \begin{align}
%   S3: f_{S_3}(x_1,x_2,x_3) &= \frac{x_1}{2}
%     \!\left(\sqrt{1+x_2+x_3^{2}}-1\right)
%     + x_1\exp\!\bigl(1+\sin x_3\bigr),\\
%   S2: f_{S_2}(x_1,x_2) &= \bigl(1+0.1\sin x_1\bigr)\,
%         f_T(x_1,x_2,0.5,0.5)-2x_1+x_2^{2}+0.75,\\
%   S1: f_{S_1}(x_1,x_2) &= x_1^{2}+\cos x_2.
% \end{align}
\begin{align*}
\mathcal S_1: f_{S_1}(\mathbf x_{S_1})
    &= x_1^{2}+\cos x_2,\\
\mathcal S_2: f_{S_2}(\mathbf x_{S_2})
    &= \bigl(1+0.1\sin x_1\bigr)\,
       f_T(x_1,x_2,0.5,0.5)\nonumber\\[-0.3ex]
       &-2x_1+x_2^{2}+0.75,\\
\mathcal S_3: f_{S_3}(\mathbf x_{S_3})
    &= \frac{x_1}{2}\Bigl(\sqrt{1+x_2+x_3^{2}}-1\Bigr)
       \nonumber\\[-0.3ex]
    &\quad+ x_1\exp\!\bigl(1+\sin x_3\bigr).
\end{align*}

In $\mathcal S_1$, the source ignores $x_3$ and $x_4$ entirely and shares no structural similarity with target; it is intentionally constructed to simulate a potentially negative source. Note that here we only intuitively identify $\mathcal S_1$—based on their functional forms—as potentially causing negative transfer. In reality, these sources may not necessarily induce negative transfer; the model will adaptively learn their relationship with the target. The same rationale applies to the designs in Simulation Case 2 and Simulation Case 3. Note that the low-fidelity data sources in this case are not associated with an inherent fidelity hierarchy. However, to enable the use of the MF-DGP-EM method, we impose an assumed ordering by treating $\mathcal S_3$ as higher fidelity than $\mathcal S_2$, and $\mathcal S_2$ as higher fidelity than $\mathcal S_1$. Moreover, since the target and sources share the input features $x_1$ and $x_2$, the DAME strategy in MGCP-R~\cite{wang2022regularized} is also applicable. As for the reference mappings derived from physical insights, since the input variables of each source are contained within those of the target, the reference mappings naturally reduce to identity mappings on the corresponding subset of target input variables. Specifically, $r_{01}(x_1,x_2,x_3,x_4) = (x_1,x_2)$, $r_{02}(x_1,x_2,x_3,x_4) = (x_1,x_2)$, $r_{03}(x_1,x_2,x_3,x_4) = (x_1,x_2,x_3)$, which also serve as the nominal mappings required in MF-DGP-EM.

For data collection, training data are sampled over $\mathcal{X}_{\mathcal T}$ and $\mathcal{X}_{\mathcal S_j}$ using Latin Hypercube Sampling (LHS), with 50 samples for the target domain and 80 samples for each source domain. The sampling range of $x_4$ in the target domain is specifically set to $[0,1.6]$ to simulate a partially observed region. All observations are corrupted by i.\,i.\,d.\ Gaussian noise with standard deviation $\sigma=0.5$. For testing, a total of $6\times6\times9\times9$ testing points are uniformly sampled over the full target domain $[0,2]^4$, where $x_1$ and $x_2$ are divided into five equal intervals, while $x_3$ and $x_4$ are divided into eight. Under these settings, the metrics at these test points measure the model’s interpolation performance (exploitation capability) within the training region and extrapolation performance (exploration capability) beyond it. 
The results reported in Table~\ref{simcase1 table} demonstrate that our method achieves the best predictive accuracy as well as the most reliable uncertainty quantification.
\begin{table*}[tb]
  \centering
  \caption{Predictive performance (mean $\pm$ std) on Simulation Case 1, over 30 LHS repetitions.}
  \label{simcase1 table}
  \begin{tabular}{lcccccc}
    \toprule
    Metric              & IMC      &MGCP-R       & MF-DGP-EM       & SA-LVGP         & TGP  & \textbf{R$^2$-HGP} \\
    \midrule
    RMSE $\downarrow$   & 6.632$\pm$3.301 & 2.983$\pm$0.727 & 3.025$\pm$1.584 & 3.114$\pm$1.714 & 6.929$\pm$2.328    & \textbf{2.401$\pm$0.612} \\
    $R^{2}$ $\uparrow$  & 0.369$\pm$0.217 & 0.939$\pm$0.056 & 0.943$\pm$0.106 & 0.907$\pm$0.094 & 0.357$\pm$0.640    & \textbf{0.969$\pm$0.043} \\
    MNLL $\downarrow$   & 15.354$\pm$13.06  &0.570$\pm$0.822  & 1.451$\pm$1.153   & 3.414$\pm$4.014   & -0.566$\pm$0.374      & \textbf{-0.745$\pm$0.344}   \\
    \bottomrule
  \end{tabular}
\end{table*}

\begin{figure}[ht!]
\centering
\includegraphics[width=\columnwidth]{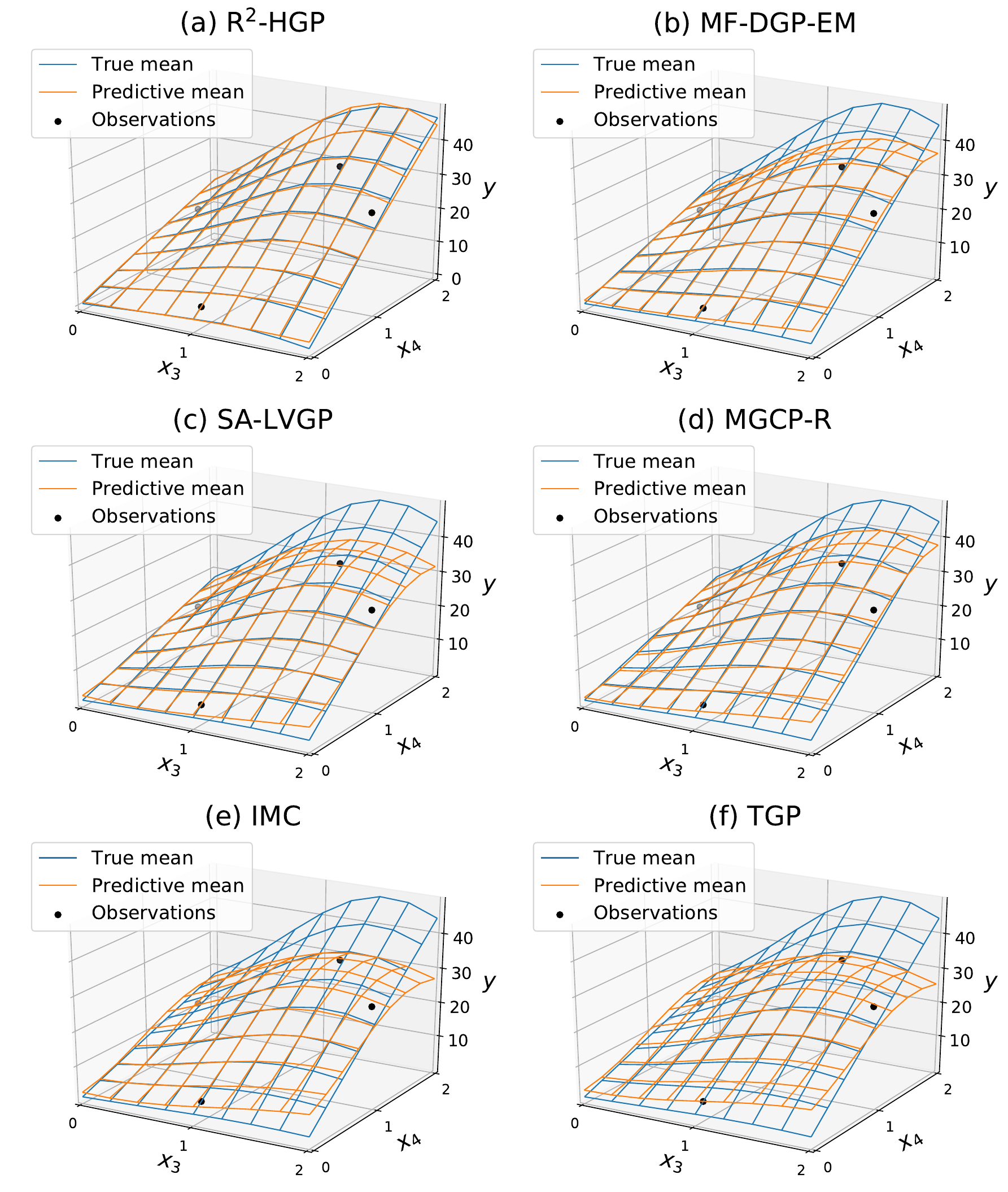}
\caption{Predictive results in one repetition of the simulation case 1 with $x_1=x_2=0.4$.}
\label{simcase1_6fig}
\end{figure}

Figure ~\ref{simcase1_6fig} shows the predicted response surface over $x_3$ and $x_4$ when $x_1 = x_2 = 0.4$ are held fixed in a single repetition. For ease of visualization, in this repetition we specifically collected four samples over $x_3$ and $x_4$ with $x_1 = x_2 = 0.4$. Since the target’s observed $x_4$ values lie only within the limited interval $[0,1.6]$, the target’s behavior for $x_4$ in the interval $[1.6,2]$ must rely on domain alignment to borrow information from other sources. The outcomes in Figure ~\ref{simcase1_6fig} demonstrate that our method achieves the best heterogeneous-domain transfer performance, providing the most accurate fit in the $x_3$–$x_4$ directions, confirming the effectiveness of our model framework.

\subsubsection{Simulation Case 2}
In this case, the features of the source input domains are not all subsets of those in the target input domain, and there exists source–target pair with no shared input features, which distinguishes this case from Simulation Case 1. Specifically, the input domains of the target and each source are:
\begin{align*}
&\mathcal{X}_{\mathcal T}:\;\mathbf x_T = (x_1,x_2)\in[0,5]^2,\\
&\mathcal{X}_{\mathcal S_1}:\;x_{S_1}=x_3\in[0,5],\\
&\mathcal{X}_{\mathcal S_2}:\;x_{S_2}=x_2\in[0,5],\\
&\mathcal{X}_{\mathcal S_3}:\;\mathbf x_{S_3}=(x_1,x_3)\in[0,5]^2.
\end{align*}
\noindent And, the target output: 
\begin{equation*}
f_T(x_1,x_2)=0.2(x_1-3)^3+0.15 x_2^2+\sin (2 x_2) .
\end{equation*}
\noindent Three source outputs:
\begin{align*}
f_{S_1}({x_3})&=0.3(x_3-3)^3,\\
f_{S_2}({x_2})&=0.3 x_2^2+2 \sin (2 x_2),\\
f_{S_3}({x_2,x_3})&=1-(x_2+x_3-4)^2.
\end{align*}

It can be observed that $f_{S_1}$ and $f_{S_3}$ both involve the feature $x_3$, which is absent from the target domain. From a physical-insight perspective, we note that $f_T$ is functionally a linear combination of a cubic term in $x_1$ and a quadratic–sinusoidal term in $x_2$. The cubic term shares the same form as $f_{S_1}$, while the quadratic–sinusoidal term matches the form of $f_{S_2}$. Accordingly, we adopt the reference mapping $r_{01}(x_1,x_2)=x_1$ for $\mathcal{X}_{\mathcal{T}}\to \mathcal{X}_{\mathcal{S}_1}$ and $r_{02}(x_1,x_2)=x_2$ for $\mathcal{X}_{\mathcal{T}}\to \mathcal{X}_{\mathcal{S}_2}$. Clearly, the order and functional form of $f_{\mathcal{S}_3}$ do not match well with those of $f_T$, suggesting that $\mathcal S_3$ may act as a weakly correlated source, which our sparsity regularizer is expected to down-weight.

In data collection, for each source \(\mathcal S_j\), we sample \(30\) training inputs from \(\mathcal{X}_{\mathcal S_j}\) using LHS. For the target, we draw \(n_{\mathcal{T}}=15\) training points using LHS within the restricted region \([0,5]\times[0,4]\subset[0,5]^2\) to emulate data scarcity in a limited area of interest. Observations are corrupted by additive Gaussian noise \(\epsilon\sim\mathcal N(0,0.2^2)\).  A separate test set of \(n_{\text{test}}=676\) inputs is collected by uniform sampling over the full plane \([0,5]\times[0,5]\) plane in \((x_1,x_2)\), with spacing 0.2 along each dimension, yielding a $26\times26$ grid. Consequently, the reported metrics reflect both interpolation error and extrapolation error beyond. 

\begin{figure}[ht!]
\centering
\includegraphics[width=\columnwidth]{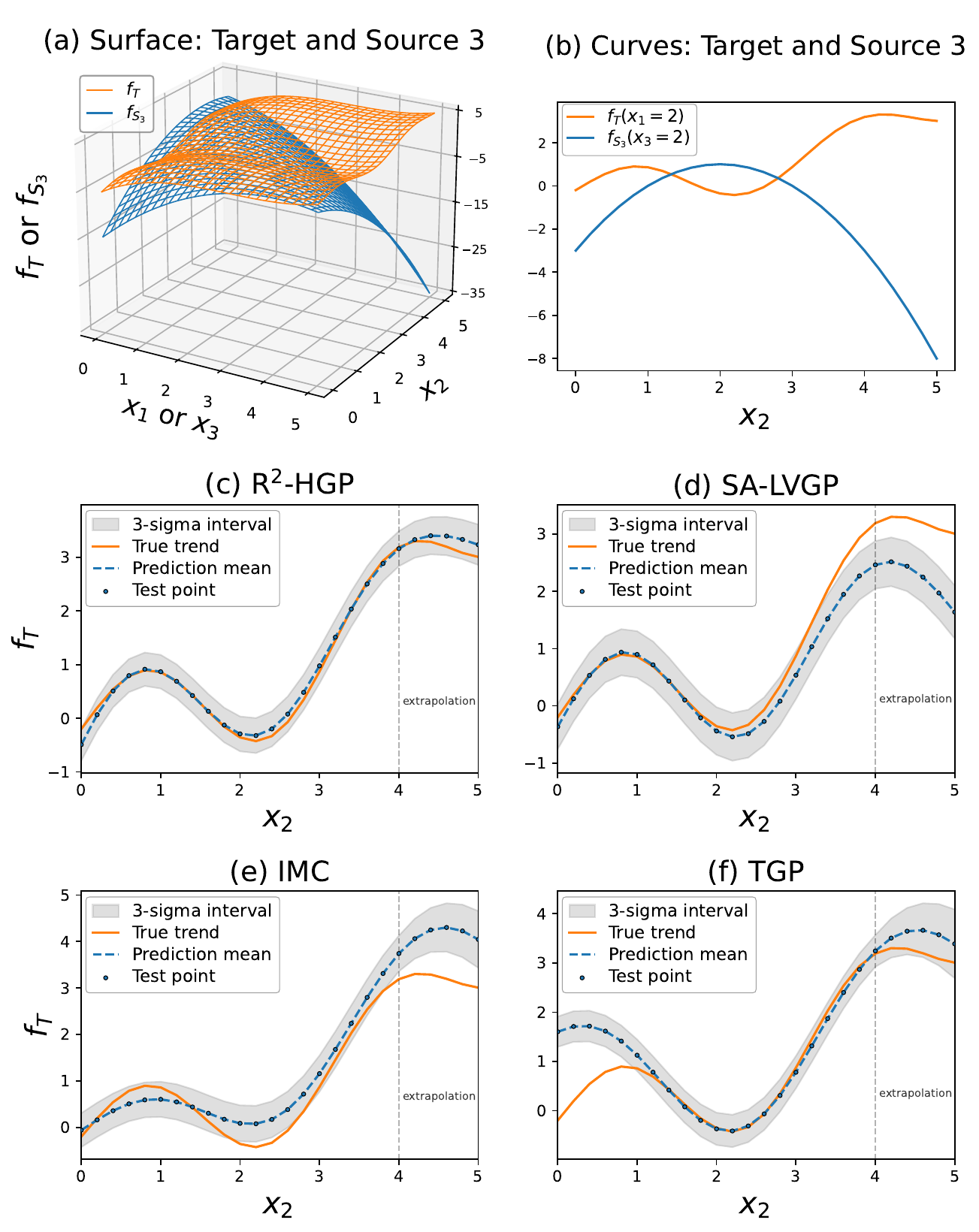}
\caption{Visualization of results in one repetition of Simulation Case 2. (a) shows the output surfaces of the target and Source 3 (based on physical insights, $x_1$ and $x_3$ are represented on the same axis); (b) presents the trend of the target and Source 3 along the $x_2$ dimension with $x_1 = x_3 = 2$; (c)--(f) illustrate the predictive performance of the target output along the $x_2$ dimension.}
\label{curvesimcase2}
\end{figure}

The experimental results are shown in Table~\ref{simcase2 table}. Fig.~\ref{curvesimcase2} shows the trend plots of the target and Source 3, along with the target’s predicted trend in a single repetition. Based on physical insights, $x_3$ should be an identity mapping of $x_1$, so in panel (a) we overlay the $x_3$ and $x_1$ axes and display the surfaces of the target and Source 3 in the same coordinate system. In panel (b), we fix both $x_3$ and $x_1$ at 2 to obtain trend plots of the target and Source 3 with respect to $x_2$. In both (a) and (b), we observe that when $x_2$ is large, the trend of Source 3 diverges significantly from that of the target. Panels (c–f) demonstrate the superiority of our method. In particular, the prediction in (d) is noticeably worse than in (f): SA-LVGP’s prediction for the target exhibits a clear downward bias in the right-hand region, indicating that the presence of Source 3 induces negative transfer. By contrast, our method remains unaffected, showing that our regularized model can effectively select informative sources. From this repetition, the learned transfer coefficients are
$\rho_1 = 0.69$, $\rho_2 = 0.48$, and $\rho_3 = 7.49 \times 10^{-4}$.
It is evident that our method successfully filtered out Source 3, which would have caused negative transfer, while at the same time capturing the correct relationships with Source 1 and Source 2.
Boxplots of the three metrics $R^2$, RMSE, and MNLL are presented in Fig.~\ref{boxplot_simcase2case3}. We observe that our R²-HGP method achieves the best performance across all metrics. From the $R^2$ and RMSE results, it is clear that SA-LVGP’s predictive performance is worse than that of TGP, which uses only the target data, confirming the occurrence of negative transfer in SA-LVGP. Furthermore, by examining MNLL, we can see that our method’s uncertainty quantification capability far surpasses that of IMC and SA-LVGP, indicating that our approach provides more accurate confidence intervals and more reliable predictive distributions. In some repetitions, SA-LVGP’s MNLL is extremely large, implying severe inaccuracy in its uncertainty modeling.
\begin{table}[t]
  \centering
  \caption{Predictive performance (mean $\pm$ std) on Simulation Case 2, over 30 LHS repetitions.}
  \label{simcase2 table}
  \begin{tabular}{lccccc}
    \toprule
    Metric              & IMC                    & SA-LVGP         & TGP  & \textbf{R$^2$-HGP} \\
    \midrule
    RMSE $\downarrow$   & 0.614$\pm$0.30  & 0.851$\pm$0.31 & 0.635$\pm$0.24    & \textbf{0.342$\pm$0.16} \\
    $R^{2}$ $\uparrow$  & 0.825$\pm$0.12  & 0.737$\pm$0.15 & 0.821$\pm$0.11    & \textbf{0.984$\pm$0.04} \\
    MNLL $\downarrow$   & 5.298$\pm$3.86      & 3.106$\pm$2.12   & \textbf{-2.143}$\pm$0.94      & -2.095\textbf{$\pm$0.73}  \\
    \bottomrule
  \end{tabular}
\end{table}
\subsubsection{Simulation Case 3}
In the third simulation case, the target input domain and source input domain are parameterized in different coordinate systems: Cartesian and spherical parametrizations, in addition to different dimensionalities. Our goal is to verify the performance of our model when the input domain is totally different. The target and source functions are defined as follows:
\begin{equation*}
\begin{aligned}
f_{T}(r,\theta,\phi)
=&3.5\,\bigl(r\cos(\tfrac{\pi}{2}\phi)\bigr)
+2.2\,\bigl(r\sin(\tfrac{\pi}{2}\theta)\bigr)\\
+&0.85\Bigl|\,r\cos(\tfrac{\pi}{2}\theta)-2r\sin(\tfrac{\pi}{2}\theta)\Bigr|^{2.2}\\
+&\frac{2\cos(\pi\phi)}{1+3r^2+10\theta^2},
\end{aligned}
\end{equation*}
\begin{equation*}
\begin{aligned}
f_{S_1}(x_1,x_2)
=&3x_1+2x_2+0.7\,\bigl|x_1-1.7x_2\bigr|^{2.35},\\
f_{S_2}(u,v)
=&\sin\bigl(5\pi u\bigr)\,e^{-v}
+0.3\,u^2
-0.4\,v^3.
\end{aligned}
\end{equation*}
\noindent The corresponding input domains are:
\begin{align*}
&\mathcal{X}_{\mathcal T}:\;\mathbf x_T = (r,\theta,\phi)\in[0,1]\times[0,\frac{\pi}{2}]^2\\
&\mathcal{X}_{\mathcal S_1}:\;\mathbf x_{S_1}=(x_1,x_2)\in[0,1]^2,\\
&\mathcal{X}_{\mathcal S_2}:\;\mathbf x_{S_2}=(u,v)\in[0,1]^2,
\end{align*}
It can be observed that $f_T$ and $f_{S_1}$ share a similar functional structure. Accordingly, the reference mapping from $\mathcal{X}_{\mathcal T}$ to $\mathcal{X}_{\mathcal S_1}$ is defined as
\begin{align*}
x &= r\cos\bigl(\tfrac{\pi}{2}\phi\bigr)\,,\\
y &= r\sin\bigl(\tfrac{\pi}{2}\theta\bigr)\,.
\end{align*}
\noindent In contrast, no explicit practical insight is available between $\mathcal S_2$ and $\mathcal T$. Moreover, the order and functional form of $f_{\mathcal S_2}$ do not match well with those of $f_T$, suggesting that $\mathcal S_2$ may serve as a weakly correlated source.

For data collection, training samples for the sources are drawn within each $\mathcal X_{\mathcal S_j}$ using LHS, with 30 samples per source domain. For the target domain, 10 samples are collected with the $r$ restricted to the interval $[0,0.8]$. For testing, $5\times10\times10$ points are uniformly sampled over $\mathcal X_{\mathcal T}$. The experimental results are shown in Table~\ref{simcase3 table}

\begin{table}[htbp]
  \centering
  \caption{Predictive performance (mean $\pm$ std) on Simulation Case 3, over 30 LHS repetitions.}
  \label{simcase3 table}
  \begin{tabular}{lccccc}
    \toprule
    Metric              & IMC                    & SA-LVGP         & TGP  & \textbf{R$^2$-HGP} \\
    \midrule
    RMSE $\downarrow$   & 0.661$\pm$0.12  & 0.702$\pm$0.14 & 0.709$\pm$0.13    & \textbf{0.384$\pm$0.09} \\
    $R^{2}$ $\uparrow$  & 0.816$\pm$0.12  & 0.754$\pm$0.12 & 0.745$\pm$0.09    & \textbf{0.949$\pm$0.05} \\
    MNLL $\downarrow$   & 10.67$\pm$6.72     & 5.78$\pm$5.51   & -1.135$\pm$\textbf{0.26}      & \textbf{-1.363}\textbf{$\pm$0.26}  \\
    \bottomrule
  \end{tabular}
\end{table}

Boxplots of the three metrics for Simulation Case 3 are presented Fig.~\ref{boxplot_simcase2case3} . It can be seen that our R²-HGP method achieves the best performance across all metrics. Notably, although IMC and SA-LVGP deliver mean-prediction performance comparable to TGP (similar $R^2$ and RMSE), their uncertainty-quantification capability is markedly inferior to TGP, as evidenced by the much larger MNLL.

\begin{figure}[ht!]
\centering
\includegraphics[width=\columnwidth]{pdf/boxplot_simcase2case3.pdf}
\caption{Boxplots of $R^2$, RMSE, MNLL in Simulation Case 2 and Simulation Case 3. The line in each box represents the median value. Note that the boxplot of Case 1 is omitted due to space limitations.}
\label{boxplot_simcase2case3}
\end{figure}

\subsection {Real Case}\label{sec:Real Case}

In this section, we evaluate R$^2$-HGP on three real-world engineering problems that exhibit distinct forms of input heterogeneity: cantilever beam design, ellipsoidal void modeling, Ti6Al4V alloy manufacturing. We aim to replicate typical research and industrial scenarios where high-quality data from the key source are scarce. The first case study illustrates a scenario where the data sources differ in input domains arising from distinct engineering designs. The second case study involves data sources that vary not only in design complexity (2D vs. 3D) but also in physical fidelity (elastic vs. plastic behavior). The third case study concerns data collected from different manufacturing modalities of the same material system, where the input spaces are entirely heterogeneous and share no common parameters. The data used for the three case studies can be found at~\cite{comlek2024heterogenous}.

\subsubsection{Real Case 1 for Cantilever Beam Design}

\begin{figure}[ht!]
\centering
\includegraphics[width=\columnwidth]{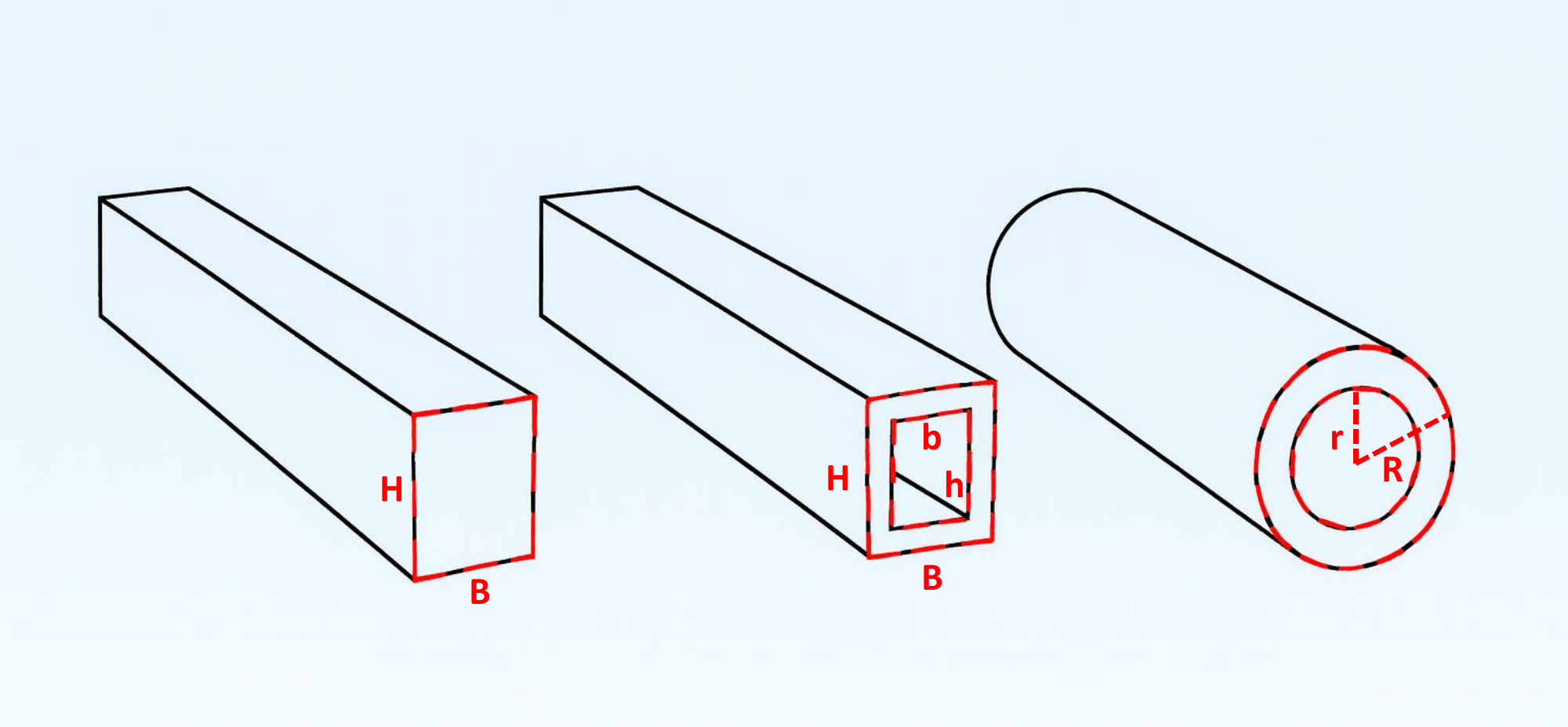}
\caption{Different cantilever beam designs with varying cross-sections.}
\label{beam}
\end{figure}

The cantilever beam design is a well-known problem of interest for many engineering applications such as bridges, towers, and buildings. A cantilever beam is a horizontally elongated structure that is fixed on one end. Typically, a structural load is applied on the unsupported end of the beam to observe its deflection and support capability. In the context of engineering design, the cantilever beam can take many shapes and forms for optimized performance. In this case, designs of the beam include Rectangular Beams (RB), Rectangular Hollow Beams (RHB), and Circular Hollow Beams (CHB), each with different design parameters and numbers of training samples, as summarized in Table~\ref{beam-sources}. Fig.~\ref{beam} shows the cross-sectional views of the three types of cantilever beams. For each of the designs, data that consists of the respective design variables and the common output, maximum deflection of the beam, is created from computer simulation by finite element modeling or analytical model under identical loading. Since CHB data is the most expensive to obtain, it has the smallest sample size and is designated as the target domain. The number of test samples for the target is 993.

\begin{table}[htbp]
  \centering
  \caption{Data description for Cantilever Beam case}
  \label{beam-sources}
  \begin{tabularx}{\linewidth}{@{}l X X c@{}}
    \toprule
    Data Type  & Input Variables & Output Variable & \#Train Samples\\
    \midrule
    RB   & $B(m),\,H(m)$ &  & 60\\ \addlinespace
    RHB  & \makecell[l]{$B(m),\,H(m),$\\$b(m),\,h(m)$} & \makecell[l]{Maximum \\Deflection (m)} & 50\\ \addlinespace
    CHB  & $R(m),\,r(m)$ &  & 15 \\
    \bottomrule
  \end{tabularx}
  \par\vspace{0.5em}
  \parbox{\linewidth}{\footnotesize
Notes: \(B\) is the outer-rectangle length, \(H\) is the outer-rectangle height, 
\(b\) is the inner-rectangle width, \(h\) is the inner-rectangle height, 
\(R\) is the outer radius, and \(r\) is the inner radius. All quantities are in meters.}

\end{table}

\subsubsection{Real Case 2 for Ellipsoidal Void Modeling}

Stress concentrations induced by voids or defects play a critical role in material failure and reliability assessment. Accordingly, the second case study investigates the modeling of 3D ellipsoidal voids, where the data sources are set up not only to vary in parametrization (as seen in the Cantilever Beam study) but also in complexity (2D versus 3D) and fidelity of the structural simulation (elastic analysis versus plastic analysis). Stress concentration around an ellipsoidal void is modeled at three levels: (1) 2D ellipse (linear elastic), (2) 3D ellipsoid (linear elastic), and (3) 3D ellipsoid with rotation (elastic–plastic), denoted as 2DE, 3DE, and 3DER, respectively. The sources vary in design complexity with additional features incorporated into the void parametrizations. Furthermore, plastic structural analysis was applied to 3DER, while 2DE and 3DE used elastic structural analysis, making 3DER distinct in simulation fidelity as well. Finite‐element analyses are used in all three cases, with increasing model complexity and nonlinearity. Consequently, the computational cost of data generation for 3DER is higher than that for 2DE and 3DE, and it is therefore designated as the target. In this case, the reference mappings from the input domain of 3DER to those of 2DE and 3DE are identity mappings on the corresponding shared features. Since the data sources are distinguished by fidelity levels and share the features $r_x(m)$ and $r_z(m)$, the methods MGCP-R and MF-DGP-EM are applicable in this case.

The details of the data domain and numbers of training samples are shown in Table~\ref{ellipse-sources}, where the Maximum von Mises Stress ($\sigma_{V_M}$) is the common output. The number of test samples for the target is 35. 

\begin{table}[htbp]
  \centering
  \caption{Data description for Ellipsoidal Void Modeling case}
  \label{ellipse-sources}
  \begin{tabular}{@{}l l l c@{}}
    \toprule
    Data Type                               & Input Variables                                     & Output Variable     &\#Train Samples             \\
    \midrule
    2DE                          & $r_x(m),\,r_z(m)$                                    &                           &200       \\\addlinespace
    3DE                          & \makecell[l]{$r_x(m),\,r_z(m)$,\\$r_y(m)$}                          & \makecell[l]{Maximum von\\Mises Stress ($\sigma_{V_M}$)} &100\\\addlinespace
    3DER        & \makecell[l]{$r_x(m),\,r_z(m),$\\$r_y(m)$,$\,\theta_y(\mathrm{degree})$} &                                  &15\\
    \bottomrule
  \end{tabular}
  \par\vspace{0.5em}
  \parbox{\linewidth}{\footnotesize
Notes: \(r_x, r_y, r_z\) are the axis lengths of the ellipsoidal void along the
X-, Y-, and Z-directions, respectively, and are measured in meters;
\(\theta_y\) is the rotation angle of the void around the Y-axis (in degrees).
}

\end{table}

\subsubsection{Real Case 3 for Ti6Al4V Alloy Manufacturing}

Titanium alloy (Ti6Al4V) has been at the forefront of aerospace applications due to its superior corrosion resistance, strength, and toughness. The manufacturing of this alloy has evolved beyond the conventional casting process, extending to additive and solid-state manufacturing processes such as Electron Beam Melting (EBM), Laser Powder Bed Fusion (LPBF), and Friction Stir Welding (FSW). In the third and final case study, a different manufacturing modalities for the same Ti6Al4V alloy is demonstrated. Specifically, three thermally driven manufacturing processes, EBM, LBPF, and FSW are selected as the data sources. A special characteristic of this study is that there is a complete non-overlap in the input parameter space of the data sources, meaning that there is no common parameter across the sources. This extreme heterogeneity tests whether our framework can map between entirely non‐overlapping input domians coming from different modalities. The data for Case 3 come from real physical experimental data, resulting in Case 3 having the smallest data volume. The data domain is depicted in Table~\ref{manufacturing-sources}, where the yield strength ($\sigma_y$) of the Ti6Al4V alloy is the output to be modeled for all three modes of manufacturing processes. Due to the lowest data availability, FSW is set to be the target. The number of test samples for the target is 5. 
\begin{table}[htbp]
  \centering
  \caption{Data description for Ti6Al4V Alloy Manufacturing case}
  \label{manufacturing-sources}
  \begin{tabular}{@{}l l l c@{}}
    \toprule
    Data Type                                          & Input Variables                                                   & Output Variable &\#Train Samples                     \\
    \midrule
    \makecell[l]{LBPF\\\cite{ref36}, \cite{LUO2023108911}} 
      & \makecell[l]{Laser Power (W),\\Laser Speed (mm/s)} &                                   &29   \\

    \addlinespace
    \makecell[l]{EBM\\\cite{ref38}} 
      & \makecell[l]{Focus Offset (mA),\\Line Offset (mm),\\Speed Factor}
      & \makecell[l]{Yield Strength\\($\sigma_y$,\,MPa)}           &10       \\

    \addlinespace
    \makecell[l]{FSW\\\cite{ref39}} 
      & \makecell[l]{Rotational Speed (rpm),\\Travel Speed (mm/min)} 
      &                           &4           \\
    \bottomrule
  \end{tabular}
\end{table}

\subsubsection{Performance Evaluation}

For each case study, we perform Monte Carlo cross-validation with 30 independent random train–test splits to account for variability arising from data partitioning. We report the mean $\pm$ one standard deviation of each metric across the thirty runs to demonstrate the robustness of our method.

\begin{figure*}[ht!]
    \centering

    % ---- 1 ----
    \begin{subfigure}{\textwidth}
        \centering
        \includegraphics[width=\textwidth]{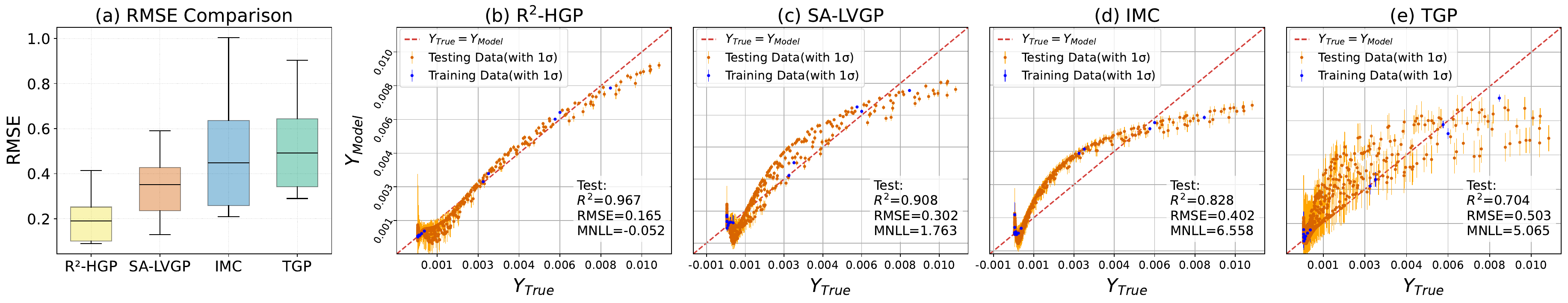}
        \caption{Boxplot of RMSE and parity plots from one repetition of Real Case 1.}
        \label{fig:realcase1}
    \end{subfigure}

    \vspace{0.42cm} % 

    % ----  2 ----
    \begin{subfigure}{\textwidth}
        \centering
        \includegraphics[width=\textwidth]{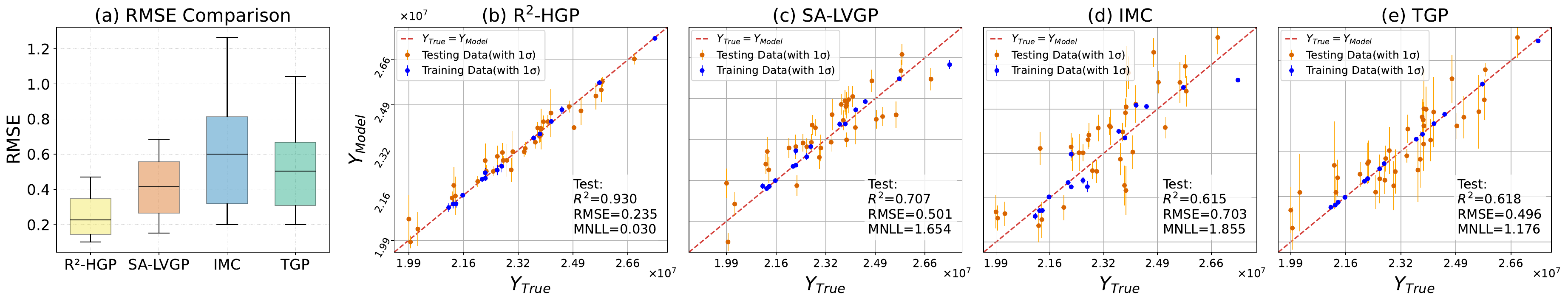}
        \caption{Boxplot of RMSE and parity plots from one repetition of Real Case 2.}
        \label{fig:realcase2}
    \end{subfigure}

    \vspace{0.42cm}

    % ----  3 ----
    \begin{subfigure}{\textwidth}
        \centering
        \includegraphics[width=\textwidth]{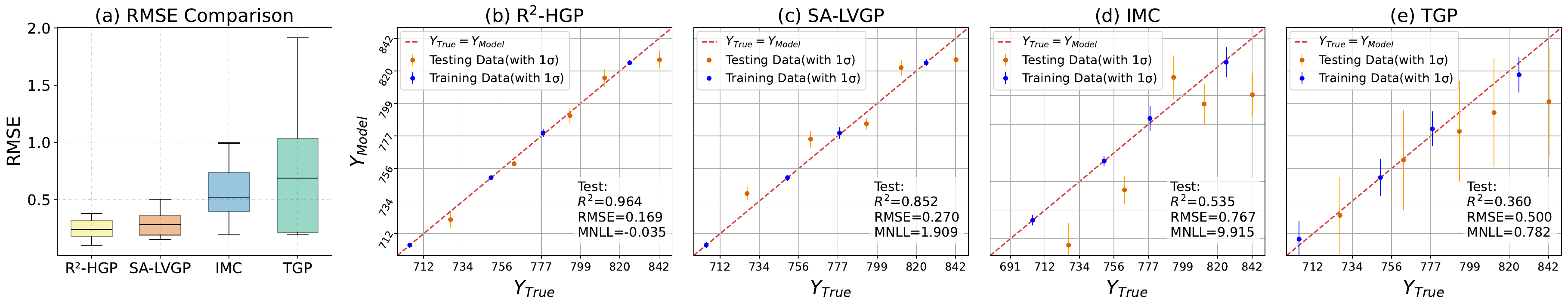}
        \caption{Boxplot of RMSE and parity plots from one repetition of Real Case 3.}
        \label{fig:realcase3}
    \end{subfigure}
    
    \captionsetup{justification=raggedright,singlelinecheck=false}
    \caption{Boxplots of RMSE and parity plots from one repetition of three real cases.}
    \label{fig:realcases}
\end{figure*}

Three evaluation metrics, $R^2$, RMSE, and MNLL, for Real Case 1, Real Case 2, and Real Case 3 are reported in Table~\ref{tab:realcase1}, Table~\ref{tab:realcase2}, and Table~\ref{tab:realcase3}, respectively. 
Figure~\ref{fig:realcase1},~\ref{fig:realcase2}, and~\ref{fig:realcase3} further illustrate RMSE boxplots and parity plots for IMC, SA-LVGP, and TGP in one representative repetition. The results show that our method achieves the best performance, while also demonstrating superior model stability and robustness.

\begin{table*}[htbp]
  \centering
  \caption{Predictive performance (mean $\pm$ std) on three real cases, computed over 30 independent runs.}
  
  \label{tab:realcases}

  % ----------------  (a) Case 1  (c) Case 3 ----------------
  \begin{minipage}[t]{\linewidth}
    \centering
    % (a) Case 1
    \begin{subtable}[t]{0.48\linewidth}
      \centering
      \caption{Real Case 1}
      \label{tab:realcase1}
      \begin{tabular}{lcccc}
        \toprule
        Metric & IMC & SA-LVGP & TGP & \textbf{R$^2$-HGP} \\
        \midrule
        RMSE $\downarrow$ & 0.451$\pm$0.25 & 0.323$\pm$0.17 & 0.530$\pm$0.19 & \textbf{0.182$\pm$0.11} \\
        $R^{2} \uparrow$  & 0.804$\pm$0.15 & 0.855$\pm$0.13 & 0.713$\pm$0.08 & \textbf{0.962$\pm$0.05} \\
        MNLL $\downarrow$ & 4.48$\pm$2.62  & 2.36$\pm$1.85  & 5.30$\pm$2.71  & \textbf{0.39$\pm$0.92}  \\
        \bottomrule
      \end{tabular}
    \end{subtable}
    \hfill
    %  (b)， Case 2
    \refstepcounter{subtable}\label{tab:realcase2}
    % (c) Case 3
    \begin{subtable}[t]{0.48\linewidth}
      \centering
      \caption{Real Case 3}
      \label{tab:realcase3}
      \begin{tabular}{lcccc}
        \toprule
        Metric & IMC & SA-LVGP & TGP & \textbf{R$^2$-HGP} \\
        \midrule
        RMSE $\downarrow$ & 0.537$\pm$0.26 & 0.283$\pm$0.13 & 0.713$\pm$1.11 & \textbf{0.235$\pm$0.09} \\
        $R^{2} \uparrow$  & 0.604$\pm$0.29 & 0.872$\pm$0.25 & 0.228$\pm$10.82 & \textbf{0.933$\pm$0.11} \\
        MNLL $\downarrow$ & 3.825$\pm$3.51 & 1.644$\pm$1.17 & 1.103$\pm$2.63 & \textbf{0.419$\pm$0.61} \\
        \bottomrule
      \end{tabular}
    \end{subtable}
  \end{minipage}

  \vspace{1.2em} % 

  % ---------------- (b) Case 2 ----------------
  \begin{minipage}[t]{\linewidth}
    \centering
    \subcaption*{(b) Real Case 2}
    \begin{tabular}{lcccccc}
      \toprule
      Metric & IMC & MGCP-R & MF-DGP-EM & SA-LVGP & TGP & \textbf{R$^2$-HGP} \\
      \midrule
      RMSE $\downarrow$ & 0.611$\pm$0.352 & 0.315$\pm$0.164 & 0.352$\pm$0.277 & 0.449$\pm$0.186 & 0.523$\pm$0.202 & \textbf{0.245$\pm$0.119} \\
      $R^{2} \uparrow$  & 0.604$\pm$0.601 & 0.861$\pm$0.117 & 0.824$\pm$0.218 & 0.714$\pm$0.121 & 0.558$\pm$0.391 & \textbf{0.933$\pm$0.107} \\
      MNLL $\downarrow$ & 1.822$\pm$1.453 & 0.564$\pm$0.670 & 1.381$\pm$1.446 & 1.524$\pm$1.375 & 0.973$\pm$0.884 & \textbf{0.325$\pm$0.549} \\
      \bottomrule
    \end{tabular}
  \end{minipage}

\end{table*}

Through above case studies, three main contributions of our method are highlighted. First, our approach demonstrates flexibility in handling diverse multi-source scenarios with varying conditions, including complexity, fidelity, and modality. Second, the predictive performance of the proposed R$^2$-HGP not only surpasses state-of-the-art baselines but also provides superior uncertainty quantification. Third, our method effectively incorporates physical insights information while remaining robust to irrelevant source domains, thereby mitigating the risk of negative transfer.

\subsection{Ablation Studies}\label{sec:Ablation Studies}

This section conducts an ablation analysis of the contributions of the two components Physical-Insight Regularization (PhyR) and Source-Selection Regularization (SSR) to quantify their independent and combined effects.

We compare four methods: HGP, the baseline model, which includes only joint variational learning of heterogeneous input alignment and multi-source transfer GP, without any regularization, with its objective function given as $\mathcal{J}_{\text{HGP}}=\mathcal{L}_{CVAE}=\mathcal{L}_{KL}+\mathcal{L}_{rec}$; PhyR-HGP, which incorporates physical-insight regularization to constrain cross-domain input relationships, with its objective function given as $\mathcal{J}_{\text{PhyR-HGP}}=\mathcal{L}_{CVAE}+\mathcal{L}_{PhyR}$; SSR-HGP, which adds an $L_1$-sparsity penalty on the transfer coefficients $\rho$, with its objective function given as $\mathcal{J}_{\text{SSR-HGP}}=\mathcal{L}_{CVAE}+\mathcal{L}_{SSR}$; and R$^2$-HGP, with its objective function given in Eq.~\eqref{eq:overal objective}. 
We report the $R^2$ and RMSE metrics for Simulation Case 2 and Real Case 1. The training setup remains consistent with the general configuration, and each method is evaluated over 30 repeated trials.

For Simulation Case 2, where one of the sources induces negative transfer, the experimental results are shown in Table ~\ref{tab:ablation-sim2}. It can be observed that PhyR-HGP brings significant improvements compared to HGP, indicating that mapping constraints based on physical insights effectively mitigate alignment biases in heterogeneous inputs. When only sparsity regularization is introduced (SSR-HGP), its performance surpasses that of PhyR-HGP, showing that in scenarios with a source that induces negative transfer, automatically removing ineffective or harmful sources through sparsification yields greater benefits than incorporating physical information alone. When both regularizations are included simultaneously (R$^2$-HGP), the performance is optimal, demonstrating the synergistic effect of physical insights providing the correct alignment direction and sparsity regularization enabling source selection.

For Real Case 1, the experimental results are shown in Table ~\ref{tab:ablation-real1}. Compared with HGP, PhyR-HGP achieves similarly stable and substantial improvements, indicating that even in the absence of an explicit practical insights, our reference mapping can still provide effective structural constraints for heterogeneous alignment. SSR-HGP shows slight improvement over HGP with a limited magnitude, and its overall performance remains lower than that of PhyR-HGP. R$^2$-HGP performs nearly on par with PhyR-HGP, suggesting that in this scenario, the sources are either generally beneficial or only very weakly harmful. Therefore, when no negative sources exist, the additional $L_1$ sparsity penalty does not adversely affect the model. Moreover, by comparing Tables~\ref{tab:ablation-sim2} and~\ref{tab:ablation-real1}, it is evident that even without any regularization terms, our HGP method still outperforms the beat baseline method IMC in Simulation case 2 and SA-LVGP in real case 1.

\begin{table}[t]
  \centering
  \caption{Predictive performance of the ablation study (mean $\pm$ std) on (a) Simulation Case 2 and (b) Real Case 1.}
  \label{tab:ablation}

  % ------- Subtable (a) Simulation Case 2 -------
  \begin{subtable}[t]{\linewidth}
    \centering
    \caption{Simulation Case 2}
    \label{tab:ablation-sim2}
    \begin{tabular}{lcccc}
      \toprule
      Metric             & HGP            & PhyR-HGP       & SSR-HGP        & \textbf{R$^2$-HGP} \\
      \midrule
      RMSE $\downarrow$  & 0.602$\pm$0.31 & 0.483$\pm$0.23 & 0.417$\pm$0.22 & \textbf{0.341$\pm$0.16} \\
      $R^2 \uparrow$     & 0.855$\pm$0.14 & 0.919$\pm$0.09 & 0.940$\pm$0.10 & \textbf{0.984$\pm$0.04} \\
      \bottomrule
    \end{tabular}
  \end{subtable}

  \vspace{0.6em} % 

  % ------- Subtable (b) Real Case 1 -------
  \begin{subtable}[t]{\linewidth}
    \centering
    \caption{Real Case 1}
    \label{tab:ablation-real1}
    \begin{tabular}{lcccc}
      \toprule
      Metric             & HGP            & PhyR-HGP       & SSR-HGP        & \textbf{R$^2$-HGP} \\
      \midrule
      RMSE $\downarrow$  & 0.278$\pm$0.13 & 0.187$\pm$0.11 & 0.268$\pm$0.15 & \textbf{0.182$\pm$0.11} \\
      $R^2 \uparrow$     & 0.891$\pm$0.11 & 0.958$\pm$0.06 & 0.902$\pm$0.11 & \textbf{0.962$\pm$0.05} \\
      \bottomrule
    \end{tabular}
  \end{subtable}

\end{table}

In summary, the ablation results clearly lead to the following conclusions: PhyR-HGP achieves significant improvements over HGP across both types of tasks, validating the critical role of our physics-insight constraints in heterogeneous input alignment. In scenarios with significant negative transfer risks (simulation case 2), further introducing the $L_1$-sparsity penalty on $\rho$ (R$^2$-HGP) yields an additional performance improvements, demonstrating the necessity of sparsity regularization for source selection and negative-transfer suppression. In Real Case 1, R$^2$-HGP performs on par with PhyR-HGP. These findings indicate that our double-regularization strategy can provide strong robustness enhancements in scenarios prone to negative transfer, while introducing no side effects in scenarios without negative transfer, thereby validating the soundness of our design.

\section{Conclusion}
\label{sec:conclu}

We propose a double‐regularized heterogeneous Gaussian process transfer model (R$^2$-HGP) to address the key challenge of input domain inconsistency between source and target domains in multi-source transfer learning. Within a variational framework, R$^2$-HGP introduces domain-alignment priors, physical-insight regularization, and source-selection regularization, enabling the model to integrate existing physical mechanisms or domain knowledge and adaptively identify and exploit relevant source information. The framework operates in an end-to-end manner, jointly optimizing the domain alignment mapping and the multi-source GP transfer model. Moreover, it exhibits strong extensibility: beyond accommodating diverse multi-source transfer scenarios, it can be readily expanded at the model level, for example, by adopting alternative kernels, incorporating different forms of regularization, or embedding various types of prior assumptions, thus enhancing its adaptability and generality. Comprehensive validation on both simulation studies and real engineering applications demonstrates that R$^2$-HGP significantly outperforms existing methods in predictive accuracy, uncertainty quantification, and robustness.

There are several open directions for future research building on this work.
The first direction concerns heterogeneity in data types. In our current study, heterogeneous inputs are restricted to the same type, namely continuous variables. However, when categorical variables or functional data variables are present, heterogeneity arises at the data-type level. In particular, for categorical inputs, appropriate kernel functions must be defined to capture covariance, which poses new challenges in heterogeneous input settings. Second, extending our framework to heterogeneous outputs is another promising direction. In many complex engineering applications, heterogeneity exists not only in inputs but also in outputs. Since heterogeneous outputs can also enable effective knowledge transfer, jointly addressing input and output heterogeneity is an important topic for future study. Finally, leveraging the strong uncertainty quantification capability of our method, the proposed R²-HGP framework can be further extended to active design tasks, such as active learning and Bayesian optimization, thereby advancing from passive prediction to data-efficient and decision-oriented modeling.

\ifCLASSOPTIONcaptionsoff
  \newpage
\fi

\bibliographystyle{IEEEtran}
\bibliography{IEEEabrv,Bibliography}

\end{document}

\maketitle

\appendices

\section{Derivation of Proposition 1}

\begin{proposition}[Validity of the Covariance]\label{prop:pd}
The covariance function constructed under the proposed framework, denoted as $\mathbf{C}_{\mathbf{Z}}(\mathbf{X},\mathbf{X})$, is guaranteed to be strictly positive definite for any choice of real value weights $\{\rho_j\}$, thereby defining a valid Gaussian-process covariance kernel.
\end{proposition}

\begin{proof}

Let the joint output vector be
\[
\mathbf{y}=(\mathbf{y}_1^\top,\mathbf{y}_2^\top,\ldots,\mathbf{y}_N^\top,\mathbf{y}_T^\top)^\top\in\mathbb{R}^n,\;n = \sum_{i=1}^{N} n_{\mathcal S_i} + n_{\mathcal{T}}.
\]
According to the proposed model,
\begin{equation*}
\begin{aligned}
    \mathbf{y}_{T}(\mathbf{x}_{T}) &= f_T(\mathbf{x}_{T}) + \epsilon_{T}(\mathbf{x}_{T}) 
    \\&= \sum_{j=1}^{N} \rho_j f_{j}(\mathbf{x}^{j}_{T}) + \delta_d(\mathbf{x}_{T})+\epsilon_{T}(\mathbf{x}_T), \\
    \mathbf{y}_{S_j}(\mathbf{x}_j) &= f_j(\mathbf{x}_j) + \epsilon_j(\mathbf{x}_j), \; j\in\mathcal{I}^S
\end{aligned}
\label{eq:transfer_model}
\end{equation*}
where $\{f_j(\cdot)\}_{j=1}^N$ and $\delta_d(\cdot)$ are mutually independent Gaussian processes, 
and $\boldsymbol{\varepsilon}=(\epsilon_i)$ are independent Gaussian noises. 
Thus, $\mathbf{y}$ can be expressed as:
\begin{equation*}
\mathbf{y}=\sum_{j=1}^N \mathbf{L}_j
\begin{bmatrix}
f_j(\mathbf{X}_j) \\
f_j(\mathbf{X}_T^j)
\end{bmatrix}
+\mathbf{L}_d\delta_d(\mathbf{X}_T)+\boldsymbol{\varepsilon}.
\end{equation*}
Here $\mathbf{L}_j \in \mathbb{R}^{\,n \times (n_{\mathcal S_j} + n_{\mathcal T})}$ and $\mathbf{L}_d\in \mathbb{R}^{\,n \times  n_{\mathcal T}}$ are block selection matrices defined as
\begin{equation*}
\mathbf{L}_j=
\begin{bmatrix}
\mathbf{0} \\
\vdots \\
\underbrace{
\begin{bmatrix}
\mathbf{I}_{n_{\mathcal S_j}} & \mathbf{0}_{n_{\mathcal S_j}\times n_{\mathcal T}}
\end{bmatrix}}_{\text{$j$-th source block}} \\
\vdots \\
\mathbf{0} \\
\underbrace{
\begin{bmatrix}
\mathbf{0}_{n_{\mathcal T}\times n_{\mathcal S_j}} & \rho_j\mathbf{I}_{n_{\mathcal T}}
\end{bmatrix}}_{\text{target block}}
\end{bmatrix},
\qquad
\mathbf{L}_d=
\begin{bmatrix}
\mathbf{0} \\
\vdots \\
\mathbf{0} \\
\mathbf{I}_{n_{\mathcal T}}
\end{bmatrix}.
\end{equation*}
Since all Gaussian components are independent, the covariance of $\mathbf{y}$ can be written as
\begin{equation*}
\begin{aligned}
\mathrm{Cov}(\mathbf{y})=&
\sum_{j=1}^N
\mathbf{L}_j
\begin{bmatrix}
\boldsymbol{K}_j(\mathbf{X}_j,\mathbf{X}_j) & \boldsymbol{K}_j(\mathbf{X}_j,\mathbf{X}_T^j) \\
\boldsymbol{K}_j(\mathbf{X}_T^j,\mathbf{X}_j) & \boldsymbol{K}_j(\mathbf{X}_T^j,\mathbf{X}_T^j)
\end{bmatrix}
\mathbf{L}_j^\top\\
&+\mathbf{L}_d\boldsymbol{K}_d(\mathbf{X}_T,\mathbf{X}_T)\mathbf{L}_d^\top+\boldsymbol{\Sigma}_{\mathrm{noise}},
\end{aligned}
\end{equation*}
where
\begin{equation*}
\boldsymbol{\Sigma}_{\mathrm{noise}}
=\mathrm{diag}(\sigma_1^2\mathbf{I}_{n_{\mathcal S_1}},\ldots,\sigma_N^2\mathbf{I}_{n_{\mathcal S_N}},\sigma_T^2\mathbf{I}_{n_{\mathcal T}}).
\end{equation*}Then, the covariance matrix $\mathrm{Cov}(\mathbf{y})$ can be expanded into the explicit block form shown in Eq.~\eqref{eq:cov-matrix}, which corresponds exactly to the covariance function defined in Eq.~(4) of the main text.

Denote
\begin{equation*}
\widetilde{\boldsymbol{K}_j}=
\begin{bmatrix}
\boldsymbol{K}_j(\mathbf{X}_j,\mathbf{X}_j) & \boldsymbol{K}_j(\mathbf{X}_j,\mathbf{X}_T^j) \\
\boldsymbol{K}_j(\mathbf{X}_T^j,\mathbf{X}_j) & \boldsymbol{K}_j(\mathbf{X}_T^j,\mathbf{X}_T^j)
\end{bmatrix}.
\end{equation*}

\begin{figure*}[!t]
  \centering
  \begin{equation}\label{eq:cov-matrix}
    \resizebox{\textwidth}{!}{%
      $\displaystyle
        \mathbf{C}_{\mathbf{Z}} (\mathbf X,\mathbf X)= 
        \left(
          \begin{array}{cccc|c}
            \boldsymbol K_1(\mathbf X_1,\mathbf X_1)+\sigma_1^2 \mathbf{I}_{n_{\mathcal S_1}}
              & 0 & \cdots & 0 
              & \rho_1 \boldsymbol{K}_1(\mathbf X_1,\mathbf X_T^1) \\
            0 
              & \boldsymbol K_2(\mathbf X_2,\mathbf X_2)+\sigma_2^2 \mathbf{I}_{n_{\mathcal S_2}}
              & \cdots & 0 
              & \rho_2 \boldsymbol{K}_2(\mathbf X_2,\mathbf X_T^2) \\
            \vdots & \vdots & \ddots & \vdots & \vdots \\
            0 & 0 & \cdots 
              & \boldsymbol K_N(\mathbf X_N,\mathbf X_N)+\sigma_N^2 \mathbf{I}_{n_{\mathcal S_N}}
              & \rho_N \boldsymbol{K}_N(\mathbf X_N,\mathbf X_T^N) \\
            \hline
            \rho_1 \boldsymbol{K}_1^T(\mathbf X_1,\mathbf X_T^1)
              & \rho_2 \boldsymbol{K}_2^T(\mathbf X_2,\mathbf X_T^2)
              & \cdots
              & \rho_N \boldsymbol{K}_N^T(\mathbf X_N,\mathbf X_T^N)
              & \begin{array}{@{}l@{}}
                  \sum_{j=1}^{N} \rho_{j}^{2}\boldsymbol{K}_j(\mathbf{X}_T^j,\mathbf{X}_T^j)\\
                  \quad{}+ \sigma_d^2\boldsymbol{K}_d(\mathbf{X}_T,\mathbf{X}_T)
                  + \sigma_T^2 \mathbf{I}_{n_{\mathcal T}}
                \end{array}
          \end{array}
        \right)$
    }
  \end{equation}
\end{figure*}

\noindent Then for any nonzero vector $\mathbf{a}\in\mathbb{R}^n$, we have
\begin{equation*}
\begin{aligned}
\mathbf{a}^\top \mathbf{C}_{\mathbf{Z}}\mathbf{a}
&=\sum_{j=1}^{N}\mathbf{a}^{\top}\mathbf{L}_{j}\widetilde{\boldsymbol{K}}_{j}\mathbf{L}_{j}^{\top}\mathbf{a}
+\mathbf{a}^{\top}\mathbf{L}_{d}\boldsymbol{K}_{d}\mathbf{L}_{d}^{\top}\mathbf{a}
+\mathbf{a}^{\top}\boldsymbol{\Sigma}_{\mathrm{noise}}\mathbf{a} \\
&=\sum_{j=1}^N \mathbf{u}_j^\top\widetilde{\boldsymbol{K}}_j \mathbf{u}_j
+\mathbf{u}_d^\top \boldsymbol{K}_d \mathbf{u}_d
+\mathbf{a}^\top\boldsymbol{\Sigma}_\mathrm{noise}\mathbf{a},
\end{aligned}
\end{equation*}
where $\mathbf{u}_j=\mathbf{L}_j^\top \mathbf{a}$ and $\mathbf{u}_d=\mathbf{L}_d^\top \mathbf{a}$. Since each kernel matrix $\widetilde{\boldsymbol{K}}_j\succeq0$ and $\boldsymbol{K}_d\succeq0$, the first two terms are
nonnegative. The noise term, however, is strictly positive:
\begin{equation*}
\mathbf{a}^\top\boldsymbol{\Sigma}_\mathrm{noise}\mathbf{a}
=\sum_{j=1}^N\sigma_j^2\|\mathbf{a}_{S_j}\|^2+\sigma_T^2\|\mathbf{a}_T\|^2>0
\qquad(\forall\,\mathbf{a}\neq0),
\end{equation*}
because $\sigma_i^2>0$ for all $i$ and at least one block of $\mathbf{a}$ is nonzero.
Therefore,
\[
\mathbf{a}^\top \mathbf{C}_{\mathbf{Z}} \mathbf{a}>0 \quad (\forall\,\mathbf{a}\neq0),
\]
which proves that $\mathbf{C}_{\mathbf{Z}}(\mathbf{X},\mathbf{X})$ is strictly positive definite. 

\end{proof}

\newpage
\section{Out-of-sample predictions}

\begin{strip}
\noindent
We here derive an approximate predictive posterior for R$^2$-HGP. Specifically, given observed data $(\mathbf{X},\mathbf{y})$, the predictive posterior for $\mathbf{f}_T(\mathbf{X}^*_T)$ at the new target inputs $\mathbf{X}^*_T$ can be obtained as shown in Eq.~\eqref{eq:prediction_final}. Note that, as the resulting distribution has no closed-form expression, we ultimately estimate it via Monte Carlo sampling. And we omit the dependency on the model and variational parameters for notational simplicity.

\begin{align}
p\bigl(\mathbf{f}_T^* \mid \mathbf{X}_T^*, \mathbf{X}, \mathbf{y}\bigr)
&= \frac{P(\mathbf{f}_T^*, \mathbf{y} \mid \mathbf{X}_T^*, \mathbf{X})}{P(\mathbf{y} \mid \mathbf{X})} \notag \\[4pt]
&= \frac{1}{P(\mathbf{y} \mid \mathbf{X})} 
\int \underbrace{P(\mathbf{f}_T^*,\mathbf{y} \mid \mathbf{Z}^*,\mathbf{Z}, \mathbf{X}_T^*, \mathbf{X})}_{\text{GP joint distribution}}
\, \underbrace{P(\mathbf{Z}^*, \mathbf{Z} \mid \mathbf{X}_T^*, \mathbf{X})}_{\text{latent variable distribution}}
\, \mathrm{d}\mathbf{Z}^*\,\mathrm{d}\mathbf{Z} \notag \\[4pt]
&= \frac{1}{P(\mathbf{y} \mid \mathbf{X})} 
\int \underbrace{p(\mathbf{f}_T^* \mid \mathbf{y},\mathbf{Z}^*,\mathbf{Z}, \mathbf{X}_T^*, \mathbf{X})}_{\text{predictive posterior}}\,\underbrace{P(\mathbf{y} \mid \mathbf{Z},  \mathbf{X})}_{\text{GP prior on training}}
\, P(\mathbf{Z}^* \mid \mathbf{X}_T^*)
\, P(\mathbf{Z} \mid \mathbf{X})
\, \mathrm{d}\mathbf{Z}^*\,\mathrm{d}\mathbf{Z} \notag \\[4pt]
&= \int 
\underbrace{p(\mathbf{f}_T^* \mid \mathbf{y},\mathbf{Z}^*,\mathbf{Z}, \mathbf{X}_T^*, \mathbf{X})}_{\text{predictive posterior}}
\, P(\mathbf{Z}^* \mid \mathbf{X}_T^*)\,\underbrace{\frac{P(\mathbf{y} \mid \mathbf{Z},\mathbf{X})\,P(\mathbf{Z} \mid \mathbf{X})}{P(\mathbf{y} \mid \mathbf{X})}}_{\text{true posterior on } \mathbf{Z}}
\, \mathrm{d}\mathbf{Z}^*\,\mathrm{d}\mathbf{Z} \notag \\[4pt]
&\approx \int 
\underbrace{p(\mathbf{f}_T^* \mid \mathbf{y},\mathbf{Z}^*,\mathbf{Z}, \mathbf{X}_T^*, \mathbf{X})}_{\text{predictive posterior}}
\, P(\mathbf{Z}^* \mid \mathbf{X}_T^*)\,\underbrace{Q(\mathbf{Z} \mid \mathbf{y},\mathbf{X})}_{\text{variational post. on} \mathbf{Z}}
\, \mathrm{d}\mathbf{Z}^*\,\mathrm{d}\mathbf{Z}\notag \\[4pt]
&\approx
\frac{1}{K\,N}
\sum_{k=1}^K
\sum_{w=1}^W
p\bigl(\mathbf{f}_T^*\mid \mathbf{y},\mathbf{Z}^{*(k)},\mathbf{Z}^{(w)},\mathbf{X}_T^*,\mathbf{X}\bigr),\,\,\mathbf{Z}^{*(k)} \sim P\bigl(\mathbf{Z}^*\mid \mathbf{X}_T^*\bigr),\,
\mathbf{Z}^{(w)} \sim Q\bigl(\mathbf{Z}\mid \mathbf{X},\mathbf{y}\bigr)
\label{eq:prediction_final}
\end{align}
\end{strip}

\section{Implementation Details}
\label{app:impl}

In this section, we provide additional technical details involved in computation, optimization and model architecture. 

In the overall loss function, the distribution regularization term $\mathcal{L}_{KL}$  aims to enforce consistency between the distributions $Q_\Phi(\mathbf{Z}\mid \mathbf{X}, \mathbf{y})$ and $P_\Theta(\mathbf{Z} | \mathbf{X})$, thereby ensuring that the prior distribution remains close to posterior distribution encoded jointly from the input–output pairs. However, our GP–based prediction requires exploiting all collected data, including the latent variables $\mathbf{Z}$ that are unobservable (see Eq. (15) and (16) in the main text). Consequently, it is necessary for the variational posterior encoder to approximate the true posterior distribution as closely as possible. Since the KL term in the standard variational lower bound(see Eq. (6) in the main text) often imposes an overly strong prior constraint, which may lead to posterior underfitting, we can introduce a hyperparameter $\beta$ (usually with $0.5<\beta<0.9$) to balance the trade-off between prior constraint and data fitting. Intuitively, reducing the KL weight shifts the optimization toward the reconstruction term, which allows the variational distribution $Q_\Phi$ more flexibility to better align with the data support, thereby yielding estimates closer to the true posterior. Accordingly, in practice, we often replace the original KL term in the original objective function with the following form:
\begin{equation}
\label{eq:KL beta}
\mathcal{L}_{KL}=\beta  \mu \Big\{ - D_{KL}\!\Big(Q_\Phi(\mathbf{Z}\mid \mathbf{X}, \mathbf{y})\,\big\|\, 
   P_\Theta(\mathbf{Z} | \mathbf{X}) \Big)\Big\},
\end{equation}
\noindent where the hyperparameters $\beta$ and $\mu$ modulate the strength of the KL penalty and its relative weight in the overall objective respectively. Tuning $\beta$ often leads to better posterior approximation and consequently improved predictive performance.

For detailed model architecture, the prior model \(P_\Theta(\mathbf Z\mid \mathbf X)\) and the recognition model
\(Q_\Phi(\mathbf Z\mid \mathbf X,\mathbf Y)\) are both modeled as fully factorized Gaussians across each source \(j\) and target sample \(i\), with the mean and variance functions parameterized by neural network architectures:
\begin{multline*}
    P_\Theta(\mathbf Z \mid \mathbf X)
=\prod_{j=1}^N\prod_{i=1}^{n_{\mathcal{T}}}
\mathcal{N}\!\bigl(\mathbf x_T^{j,(i)};\,g_{\mu}(\mathbf{x}_T^{(i)};\theta_j),\Sigma_{\sigma}(\mathbf{x}_T^{(i)};\theta_j)\bigl),
\end{multline*}
\begin{multline}
Q_\Phi(\mathbf Z \mid \mathbf X,\mathbf Y)
=\prod_{j=1}^N\prod_{i=1}^{n_{\mathcal{T}}}
\mathcal{N}\!\bigl(\mathbf x_T^{j,(i)};\,\mu_{T,\Phi}^{j,(i)}(\mathbf X,\mathbf Y),\\\mathrm{diag}\bigl((\sigma_{T,\Phi}^{j,(i)}(\mathbf X,\mathbf Y))^2\bigr)\bigr).
\label{eq:variational distri}
\end{multline}

\noindent The two models differ in their inputs, where the prior network takes only $\mathbf{x}_T^{(i)}$ as input for each source $j$:
\begin{align*}
g_{\mu}(\mathbf{x}_T^{(i)};\theta_j)
&= \psi\!\bigl(A_{j,1}\,\mathbf{x}_T^{(i)} + b_{j,1}\bigr),\\
\Sigma_{\sigma}(\mathbf{x}_T^{(i)};\theta_j)
&= \Sigma\!\bigl(A_{j,2}\,\mathbf{x}_T^{(i)} + b_{j,2}\bigr),
\end{align*}
Here \(\psi(\cdot)\) denotes a generic activation function and $\Sigma(\cdot)$ governs the structure of the covariance.

In principle, the posterior of $\mathbf{x}_T^{j,(i)}$ should depend on all observed data and be mutually coupled. To facilitate computation, we make a tractable approximation as shown in Eq.~\eqref{eq:variational distri}, and constructing the input to the recognition network as a composite vector that summarizes both the target-domain data and the corresponding source-domain-$j$ data. Specifically, for source $j$ and target sample $i$, the input to the recognition network is defined as:
\begin{equation*}
\begin{split}
&\mathbf{u}_j^{(i)}(\mathbf X,\mathbf Y)
= \\
&T\Bigl(
    \mathbf{x}_T^{(i)},\,y_T^{(i)},
    \underbrace{\bar{\mathbf{x}}_T}_{\frac1{n_{\mathcal{T}}}\sum_k\mathbf{x}_T^{(k)}},
    \underbrace{\bar{y}_T}_{\frac1{n_{\mathcal{T}}}\sum_k y_T^{(k)}},
    \underbrace{\alpha_j\bar{\mathbf{x}}_j}_{\frac{\alpha_j}{n_{\mathcal{S}_j}}\sum_k\mathbf{x}_j^{(k)}},
    \underbrace{\alpha_j\bar{y}_j}_{\frac{\alpha_j}{n_{\mathcal{S}_j}}\sum_k y_j^{(k)}}
  \Bigr)
\end{split}
\end{equation*}

\noindent where $T(\cdot)\in\mathbb{R}^d$ denotes a concatenation operator. The vector $\mathbf{u}_j^{(i)}$ aggregates information from the current target-domain sample $(\mathbf{x}_T^{(i)}, y_T^{(i)})$, the mean statistics of the target domain $(\bar{\mathbf{x}}_T, \bar{y}_T)$, and the mean statistics of the source domain $(\bar{\mathbf{x}}_j, \bar{y}_j)$ into a single representation that summarizes both target- and source-domain data. In our implementation, coefficients $\alpha_j$ are introduced to control the relative contributions of target-domain and source-domain statistics, thereby adjusting their influence on posterior inference. The recognition network then takes $\mathbf{u}_j^{(i)}$ as input:
\begin{align*}
    h_{\Phi}^{\,j,(i)}
\;&=\;
\psi\!\bigl(A_{1}'\,\mathbf{u}_j^{(i)} + b_{1}'\bigr),\\
\quad
\mu_{T,\Phi}^{\,j,(i)}
\;&=\;
A_{2}'\,h_{\Phi}^{\,j,(i)} + b_{2}',\\
\quad
\log\!\bigl((\sigma_{T,\Phi}^{\,j,(i)})^2\bigr)
\;&=\;
A_{3}'\,h_{\Phi}^{\,j,(i)} + b_{3}'.
\end{align*}

\noindent With these definitions, we can perform efficient sampling and closed-form KL-divergence computation for both the prior model \(P_{\Theta}(\mathbf{Z}\mid \mathbf{X})\) and the recognition model \(Q_{\Phi}(\mathbf{Z}\mid \mathbf{X},\mathbf{Y})\). Specifically, the KL divergence in Eq.~\eqref{eq:KL beta} can decompose across sources \(j\) and target samples \(i\) as
\begin{equation*}
\begin{split}
  &D_{KL}\bigl(Q_\Phi(\mathbf Z\mid \mathbf X,\mathbf Y)\;\|\;P_\Theta(\mathbf Z\mid \mathbf X)\bigr)\\
  &\quad
   =\sum_{j=1}^N\sum_{i=1}^{n_{\mathcal T}}
     D_{KL}\Bigl(
       \mathcal{N}\bigl(
         \mu_{T,\Phi}^{j,(i)},\,\mathrm{diag}\bigl((\sigma_{T,\Phi}^{j,(i)})^2\bigr)
       \bigr)\\
  &\quad\quad\quad\quad
      \big\|\;
       \mathcal{N}\bigl(
         g_{\mu}(\mathbf{x}_T^{(i)};\theta_j),\,
         \Sigma_{\sigma}(\mathbf{x}_T^{(i)};\theta_j)
       \bigr)
     \Bigr)\!.
\end{split}
\end{equation*}

\noindent When $\Sigma_{\sigma}(\mathbf{x}_T^{(i)};\theta_j)$ reduces to the diagonal form $\mathrm{diag}\bigl((g_{\sigma}(\mathbf{x}_T^{(i)};\theta_j))^2\bigr)$, the explicit element-wise closed-form expression for each pair $(j,i)$ can be given by: 
\begin{equation*}
\begin{split}
  &D_{KL}\!\Bigl(\mathcal{N}(\mu_{T,\Phi}^{j,(i)},\,\mathrm{diag}((\sigma_{T,\Phi}^{j,(i)})^2))\;\\&\big\|\;
       \mathcal{N}(g_{\mu}(\mathbf{x}_T^{(i)};\theta_j),\,\mathrm{diag}((g_{\sigma}(\mathbf{x}_T^{(i)};\theta_j))^2))\Bigr)\\
  &= \frac12\sum_{k=1}^{d_{\mathcal{S}_j}}
       \Biggl[
         \ln\frac{\bigl(g_{\sigma}(\mathbf{x}_T^{(i)};\theta_j)\bigr)_k^2}
                  {\bigl(\sigma_{T,\Phi}^{j,(i)}\bigr)_k^2}
         -1
         +\frac{\bigl(\sigma_{T,\Phi}^{j,(i)}\bigr)_k^2}
                 {\bigl(g_{\sigma}(\mathbf{x}_T^{(i)};\theta_j)\bigr)_k^2}
  \Biggr.\\
  &\quad\Bigl.
         +\,\frac{\bigl(\mu_{T,\Phi}^{j,(i)}-g_{\mu}(\mathbf{x}_T^{(i)};\theta_j)\bigr)_k^2}
                 {\bigl(g_{\sigma}(\mathbf{x}_T^{(i)};\theta_j)\bigr)_k^2}
       \Biggr].
\end{split}
\end{equation*}

\noindent For reparameterization trick in the reconstruction term ${\mathcal{L}}_{rec}$, we generate Monte Carlo samples from both the prior $P_{\Theta}(\mathbf{Z}\mid \mathbf{X})$ and the variational posterior \(Q_{\Phi}(\mathbf Z\mid \mathbf X,\mathbf Y)\). For each source \(j\), target‐sample index \(i\), and draw index \(l=1,\dots,L\), we first sample $\varepsilon^{\,j,(i),(l)} \;\sim\;\mathcal{N}\bigl(\mathbf{0},\,\mathbf{I}\bigr)$. The corresponding latent variable sample is then obtained as:
\[
\mathbf{x}_T^{\,j,(i),(l)}
=
h_{\Phi}^Q\bigl(\varepsilon^{\,j,(i),(l)},\mathbf X,\mathbf Y\bigr)
=
\mu_{T,\phi}^{\,j,(i)}
\;+\;
\sigma_{T,\phi}^{\,j,(i)} \;\odot\;\varepsilon^{\,j,(i),(l)}.
\]
\noindent Collecting all these per‐source, per‐sample draws yields the \(l\)-th Monte Carlo sample of the variational posterior:
\[
\mathbf Z^{(l)}
=\bigl\{\,
\mathbf{x}_T^{\,j,(i),(l)} : j=1,\dots,N,\quad i=1,\dots,n_T
\bigr\}.
\]
\noindent Analogously, for the prior we obtain samples $\mathbf Z^{(m)}$ by 
\begin{align*}
    \mathbf{x}_T^{\,j,(i),(m)}
&=
h_{\Theta}^P\bigl(\varepsilon^{\,j,(i),(m)},\mathbf{x}_T)\\
&=
g_{\mu}(\mathbf{x}_T^{(i)};\theta_j)
+
g_{\sigma}(\mathbf{x}_T^{(i)};\theta_j)\odot\varepsilon^{\,j,(i),(m)}.
\end{align*}
\noindent The two sets of samples \(\{\mathbf Z^{(l)}\}_{l=1}^L\) and \(\{\mathbf Z^{(m)}\}_{m=1}^M\) are subsequently used in Eq.~\eqref{eq:loss rec}. Accordingly, our GP-based decoder’s reconstruction term can be formulated as:
\begin{equation}
\begin{aligned}
{\mathcal{L}}_{rec} 
&= -\frac{\mu}{2L}\sum_{l=1}^L 
    \left[
\mathbf{y}^{\top} \boldsymbol{C}_{Z^{(l)}}^{-1} \mathbf{y}
+ \log\det(\boldsymbol{C}_{\mathbf{Z}^{(l)}})
+ n \log(2\pi)
\right]\\
&\quad -(1-\mu)\,\frac{1}{2M}\sum_{m=1}^M 
    \log \Big[ 
\mathbf{y}^{\top} \boldsymbol{C}_{\mathbf{Z}^{(m)}}^{-1} \mathbf{y}
+ \log\det(\boldsymbol{C}_{\mathbf{Z}^{(m)}})\\
&\quad+ n \log(2\pi)
\Big].
\end{aligned}
\label{eq:loss rec}
\end{equation}

Additionally, it is worthy noting that our physical-insight regularization term does not necessarily require the reference mapping $g_0(\cdot)$ to exist. Instead, the reference-mapped values $g_0(\mathbf{X}_T)$ of the observed points are sufficient to construct the physical-insight regularization term.

\begin{figure*}[t] % 
    \centering
    \begin{tikzpicture}[
      node distance=5mm and 10mm,
      >=Latex,
      every node/.style={font=\small,inner sep=6pt},
      startstop/.style={rounded rectangle, draw, align=center, minimum width=30mm, minimum height=6mm},
      process/.style={rectangle, draw, align=left, minimum width=40mm, minimum height=7mm},
      decision/.style={diamond, draw, aspect=2, align=center, inner sep=1pt},
      io/.style={trapezium, trapezium left angle=70, trapezium right angle=110, draw, align=center, minimum width=42mm, minimum height=7mm}
    ]
    \node[startstop] (start) {Start};
    \node[io, below=of start] (input) {Input: \\ Multiple sources $\{(\mathbf{X}_i,\mathbf{y}_i)\}_{i=1}^N$, target $(\mathbf{X}_T,\mathbf{y}_T)$};
    \node[process, below=of input] (loop) {For each source–target pair $j=1,\dots,N$: collect physical insight};
    \node[decision, below=of loop] (dec) {Explicit \\ physical insight?};
    \node[process, below left=5mm and 10mm of dec] (nom) {Obtain nominal mapping $g_{0j}(\cdot)$ \\or nominal mapped values $g_{0j}(\mathbf{X}_T)$};
    \node[process, below right=5mm and 15mm of dec] (imc) {Apply reference mapping $r_{0j}^{\text{IMC}}(\mathbf{X}_T)$ \\for physical insight extraction};
    \node[process, below=15mm of dec] (collect) {Aggregate $\{r_{0j}(\mathbf{X}_T)\}$ by taking $g_{0j}$ or $r_{0j}^{\text{IMC}}$};
    
    \node[process, below=of collect] (init) {Initialize prior $p_\Theta(\mathbf{Z}\mid \mathbf{X})$, \\encoder $q_\Phi(\mathbf{Z}\mid \mathbf{X},\mathbf{y})$,\\and decoder $P_\Theta(\mathbf{y}\mid \mathbf{X},\mathbf{Z})$};

    \node[process, below=of init] (train) {
    
    Compute $\mathcal{L}_{\text{rec}}$ from Eq.~\eqref{eq:loss rec}, $\mathcal{L}_{\text{KL}}$ from Eq.~\eqref{eq:KL beta},\\
    
    maximize $\mathcal{L}_{KL}+\mathcal{L}_{rec}+\mathcal{L}_{PhyR}+\mathcal{L}_{SSR}$ 
    
    };

    \node[process, below=of train] (predict) {Prediction for new input $\mathbf{X}_T^\ast$: \\ sample $\mathbf{Z}^\ast \sim P_{\widehat{\Theta}}$ and $\mathbf{Z} \sim Q_{\widehat{\Phi}}$,\\ use Eq.~\eqref{eq:prediction_final} for prediction};
    \node[startstop, below=of predict] (end) {End};
    
    \draw[->] (start) -- (input);
    \draw[->] (input) -- (loop);
    \draw[->] (loop) -- (dec);
    \draw[->] (dec) -- node[above left]{Yes} (nom);
    \draw[->] (dec) -- node[above right]{No} (imc);
    \draw[->] (nom.south) |- (collect.west);
    \draw[->] (imc.south) |- (collect.east);
    \draw[->] (collect) -- (init);
    \draw[->] (init) -- (train);
    \draw[->] (train) -- (predict);
    \draw[->] (predict) -- (end);
    \end{tikzpicture}
    \caption{Flowchart of the proposed R$^2$-HGP method.}
    \label{fig:flowchart}
\end{figure*}
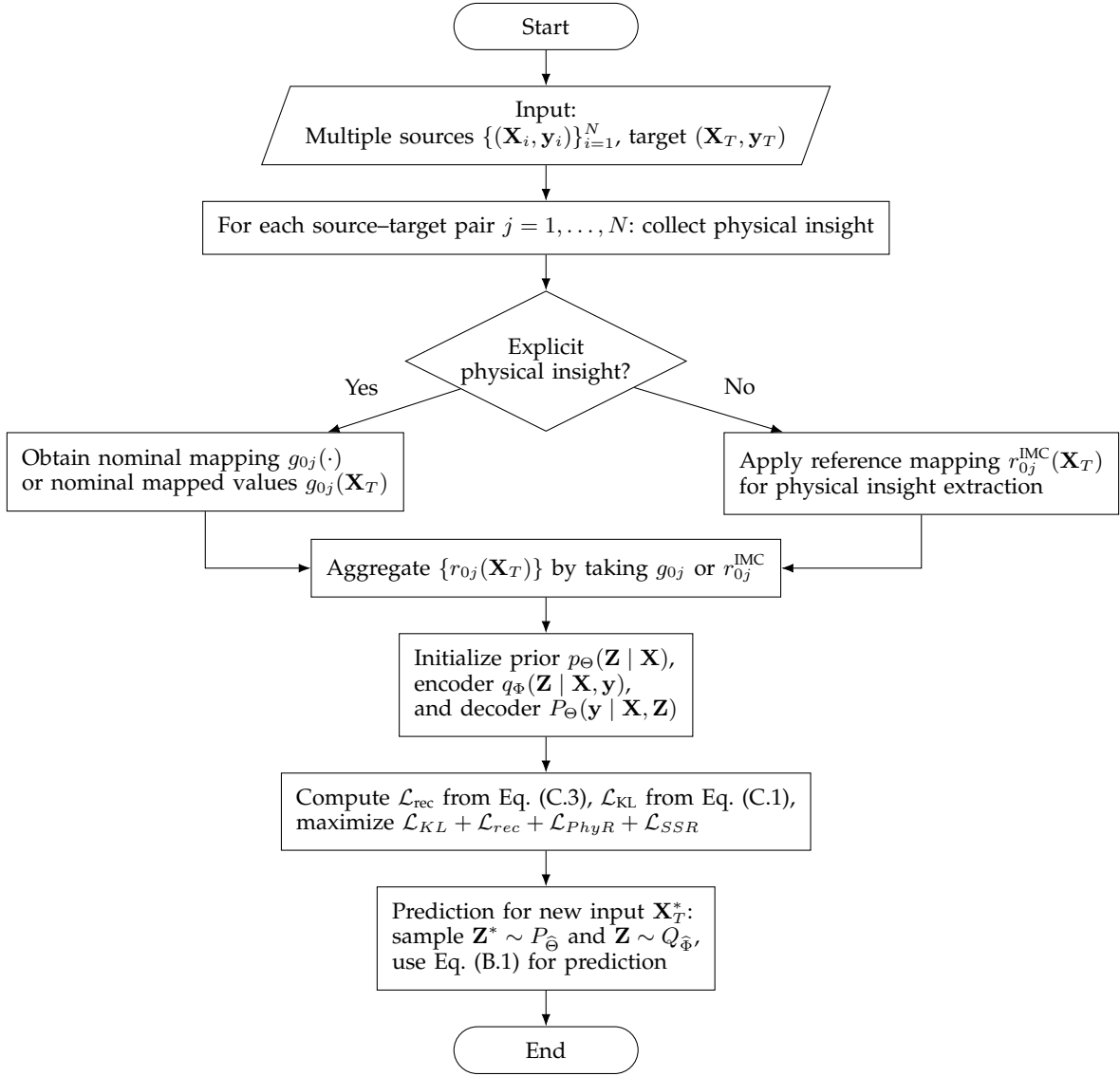

\noindent Therefore, a detailed flowchart of the complete workflow is provided in Fig.~\ref{fig:flowchart}

\section{Experiments}
In this section, we detail the experimental setups and evaluation metrics.

We evaluate all benchmark methods on the test set $\{\mathbf x_T^{*(i)},\,y_T(\mathbf x_T^{*(i)})\}_{i=1}^{n_{\text{test}}}$. For a test input $\mathbf x^*_T$, let 
$\widehat f_T(\mathbf x^*_T)$ denote the predictive mean and
$\widehat\sigma_T^{2}(\mathbf x^*_T)$ the corresponding predictive variance. For the entire set of test inputs $\mathbf X_T^*$, let
\[
  \widehat p
  \bigl(\mathbf f\mid\mathbf X_T^*\bigr)
  \;=\;
  \mathcal N\!\Bigl(
    \mathbf f;\,
    \widehat{\boldsymbol\mu}_T(\mathbf X_T^*),\,
    \widehat{\mathbf V}_T(\mathbf X_T^*)
  \Bigr)
\]
denote the predictive distribution of the corresponding test outputs. Here,  
\(\widehat{\boldsymbol\mu}_T(\mathbf X_T^*)\in\mathbb R^{n_{\text{test}}}\)  
is the joint predictive mean vector, and  
\(\widehat{\mathbf V}_T(\mathbf X_T^*)\in\mathbb R^{n_{\text{test}}\times n_{\text{test}}}\)  
is the associated covariance matrix. The ground-truth outputs are assembled in the vector
\[
\mathbf y_T^{\text{test}}
  =\bigl[\,y_T(\mathbf x_T^{*(1)}),\,\dots,\,y_T(\mathbf x_T^{*(n_{\text{test}})})\,\bigr]^{\!\top}.
\]
\noindent We adopt three complementary criteria:

\noindent\textbf{Root-Mean-Squared Error (RMSE):}
\[
  \mathrm{RMSE}
    =\sqrt{\frac{1}{n_{\mathrm{test}}}
            \sum_{i=1}^{n_{\mathrm{test}}}
            \bigl(\widehat f_T(\mathbf x_T^{*(i)})-y_T(\mathbf x_T^{*(i)})\bigr)^2}\,.
\]

\noindent\textbf{Coefficient of Determination (\textbf{$\boldsymbol{R}^2$}
):}
\[
R^{2}
=
1-
\frac{
  \displaystyle
  \sum_{i=1}^{n_{\text{test}}}
  \bigl(
    \widehat f_T(\mathbf x_T^{*(i)})-
    y_T(\mathbf x_T^{*(i)})
  \bigr)^2
}{
  \displaystyle
  \sum_{i=1}^{n_{\text{test}}}
  \bigl(
    y_T(\mathbf x_T^{*(i)})-\bar y_T
  \bigr)^2
},\,
\bar y_T=\frac{\sum_{i=1}^{n_{\text{test}}}y_T(\mathbf x_T^{*(i)})}{n_{\text{test}}}   
\]

\noindent\textbf{Mean Negative Log-Likelihood (MNLL):}\\
\begin{IEEEeqnarray*}{rCl}
  \mathrm{MNLL}
    &=& -\frac{1}{n_{\mathrm{test}}}
          \log \widehat p_T\bigl(\mathbf y_T^{\mathrm{test}}\mid\mathbf X_T^*\bigr)
       \\[0.8ex]
    &=& \tfrac{1}{2\,n_{\mathrm{test}}}\Bigl[
           (\mathbf y_T^{\mathrm{test}}-\widehat{\boldsymbol\mu}_T)^{\!\top}
           \widehat{\mathbf V}_T^{-1}
           (\mathbf y_T^{\mathrm{test}}-\widehat{\boldsymbol\mu}_T)
         \Bigr.
       \\[0.8ex]
    &&\quad\Bigl.
           +\,\log\det\widehat{\mathbf V}_T
           +\,n_{\mathrm{test}}\log(2\pi)
         \Bigr].
\end{IEEEeqnarray*}

\noindent\textbf{RMSE} quantifies the average deviation between predicted and observed values; smaller values indicate higher predictive accuracy.  
\textbf{$\boldsymbol{R}^2$}
measures the proportion of variance in the data explained by the model; values closer to~1 imply a better fit.  
\textbf{MNLL} rewards models that assign high probability to the true targets while appropriately reflecting predictive variance. In the special case where conditional independence across test inputs is assumed, \(\widehat{\mathbf V}_T\) reduces to a diagonal matrix, and the joint MNLL simplifies to the mean of point-wise MNLLs, coinciding with the standard per-sample definition.

In the simulation cases, we always design one source that may cause negative transfer for each target, along with several sources that are similar to the target. Take Simulation Case 1 as an example, in $\mathcal S_3$ we eliminate variation in the fourth target coordinate:
the $x_4/x_1^2$ factor inside the square root is set to constant 1 and the $(x_1+3x_4)$ prefactor in the exponential term is collapsed to $x_1$. Thus the response no longer depends on $x_4$, but the $(x_2+x_3^2)$ structure and the $\exp(1+\sin x_3)$ behavior are preserved. In $\mathcal S_2$, the source evaluates the target at the constant values $x_3=x_4=0.5$ (i.e., $x_3$ and $x_4$ are held fixed and not treated as inputs), and then multiplied a small sinusoidal scaling term $(1+\sin x_1/10)$, and adds a bias polynomial.
In $\mathcal S_1$, the source ignores $x_3$ and $x_4$ entirely and shares no structural similarity with target's $(x_2+ x_3^2)$ or exponential terms; it is intentionally constructed to simulate a source that may be weakly (or even negatively) transferable. The design principles for the remaining cases are similar.

Additionally, for the IMC method used as a baseline, Table~\ref{imc_simu},~\ref{imc_real} summarizes the selection of transfer sources adopted for information transfer across all experiments.

\begin{table}[!ht]
\centering
\caption{Transfer source selection for the IMC method (Simulation Cases)}
\begin{tabular}{lccc}
\hline
\textbf{Case ID} & SIMU 1 & SIMU 2 & SIMU 3 \\ 
\hline
\textbf{Transfer Source} & $\mathcal{S}_3$ & $\mathcal{S}_1$ & $\mathcal{S}_1$ \\ 
\hline
\end{tabular}
\label{imc_simu}
\end{table}

\begin{table}[!ht]
\centering
\caption{Transfer source selection for the IMC method (Real Cases)}
\begin{tabular}{lccc}
\hline
\textbf{Case ID} & REAL 1 & REAL 2 & REAL 3 \\ 
\hline
\textbf{Transfer Source} & $\mathcal{S}_1$ & $\mathcal{S}_2$ & $\mathcal{S}_1$ \\ 
\hline
\end{tabular}
\label{imc_real}
\end{table}

\noindent In each case, one source domain is designated for the baseline IMC method, serving as the alignment basis.\\\\\\\\\\\\\\\\\\\\\\\\\\\\\\\\\\\\\\\\\\\\\\\\\\\\\\\\\\\\\\\\\\\\\\\\\\\\\\\\\\\\\\\\